\documentclass{article}

    \PassOptionsToPackage{numbers, compress}{natbib}
    \usepackage[preprint]{neurips_2022}

\usepackage[utf8]{inputenc} % allow utf-8 input
\usepackage[T1]{fontenc}    % use 8-bit T1 fonts
\usepackage{hyperref}       % hyperlinks
\usepackage{url}            % simple URL typesetting
\usepackage{booktabs}       % professional-quality tables
\usepackage{amsfonts}       % blackboard math symbols
\usepackage{nicefrac}       % compact symbols for 1/2, etc.
\usepackage{microtype}      % microtypography
\usepackage{xcolor}  
\usepackage{graphicx}
\usepackage{pifont}
\newcommand{\cmark}{\ding{51}}%
\newcommand{\xmark}{\ding{55}}%
\usepackage{adjustbox}
\usepackage{multirow}

\title{VLMbench: A Compositional Benchmark for \\ Vision-and-Language Manipulation
}

\author{%
  Kaizhi Zheng%\thanks{Project Website: \url{https://sites.google.com/ucsc.edu/vlmbench}} 
  \\
  University of California, Santa Cruz\\
  \texttt{kzheng31@ucsc.edu} \\
   \And
   Xiaotong Chen \\
   University of Michigan, Ann Arbor \\
   \texttt{cxt@umich.edu} \\
   \AND
   Odest Chadwicke Jenkins \\
   University of Michigan, Ann Arbor \\
   \texttt{ocj@umich.edu} \\
   \And
   Xin Eric Wang \\
   University of California, Santa Cruz \\
   \texttt{xwang366@ucsc.edu} \\
}

\begin{document}

\maketitle

\begin{abstract}
%   One crucial ability of embodied agents is to finish tasks by following language instructions. Language can represent complicated tasks and distinguish their differences, and it is natural for humans to use language to command an embodied agent. 
%   \xin{The first two sentences have little to do with manipulation. The connection is not straightforward.}
Benefiting from language flexibility and compositionality, humans naturally intend to use language to command an embodied agent for complex tasks such as navigation and object manipulation. In this work, we aim to fill the blank of the last mile of embodied agents---object manipulation by following human guidance, e.g., “move the red mug next to the box while keeping it upright.” To this end, we introduce an Automatic Manipulation Solver (AMSolver) system and build a Vision-and-Language Manipulation benchmark (VLMbench) based on it, containing various language instructions on categorized robotic manipulation tasks. Specifically, modular rule-based task templates are created to automatically generate robot demonstrations with language instructions, consisting of diverse object shapes and appearances, action types, and motion constraints. We also develop a keypoint-based model 6D-CLIPort to deal with multi-view observations and language input and output a sequence of 6 degrees of freedom (DoF) actions. We hope the new simulator and benchmark will facilitate future research on language-guided robotic manipulation.
\end{abstract}

\section{Introduction}
\label{sec:intro}
% \xin{The intro is a bit lengthy and too technical.}
``Can you help me to clean the disks in the sink?" --- humans communicate with each other using language to issue tasks and specify the requirements.
% Humans are naturally intended to use language to convey the request for tasks requiring complicated perception processing, appropriate natural language understanding, and accurate manipulation execution. 
Although recent progress in embodied AI pushes intelligent robotic systems to reality closer than any other time before, it is still an open question how the agent learn to manipulate objects following language instructions.
% \xin{there is existing work for VLM so why is this still an open question? Unclear to me what the novelty is here.} 
Therefore, we introduce the task of Vision-and-Language Manipulation (VLM), where the agent is required to follow language instructions to do robotic manipulation.
There are recent benchmarks developed to evaluate robotic manipulation tasks with language guidance and visual input~\cite{james2020rlbench,ahmed2020causalworld,yu2020meta}. However, the collected task demonstrations are not modular and can hardly scale because they lack (1) adaptation to novel objects (2) categorization for modular and flexible composition to complex tasks. Additionally, the lack of variations in language also lead to biases for visual reasoning learning. To deal with these problems, we expect an inclusive, modular, and scalable benchmark to evaluate embodied agents for various language-guided manipulation tasks.
% \xin{I am not sure what you mean here... In general, it is unclear to me how is our benchmark different from previous language-instruction manipulation benchmarks, which we should definitely address}
% accomplish various daily tasks with novel objects in unstructured environments in the future reality. 
% In order to make a step forward, it is essential to propose a VLM benchmark.

% Recent progress in embodied AI pushes intelligent robotic systems to reality closer than any other time before. To achieve the final goal of interacting with unstructured environments to accomplish various daily tasks, the agent needs to learn how to manipulate objects through visual observations and natural languages appropriately. Compared with vision-only systems, natural language possesses two essential properties that enhance robot manipulation task learning: compact and flexible specification of various tasks and natural interactive communication interface with humans.
% We name such an agent that learns from the combined knowledge of language and vision to accomplish robot manipulation tasks a \textbf{Vision-and-Language Manipulation (VLM)} embodied agent.
% Given the rapid growth in VLM research and their limitations on generalization across different tasks, we expect an inclusive, modular, and scalable benchmark to evaluate embodied agents for various manipulation tasks. \xin{I think we need to introduce the VLM task from a broader view. Why VLM (not just technically)? What are the real world applications? The ECCV reviewers are quite confused about this.}

\begin{figure}[!tbp]
% \flushright
\includegraphics[width=\columnwidth]{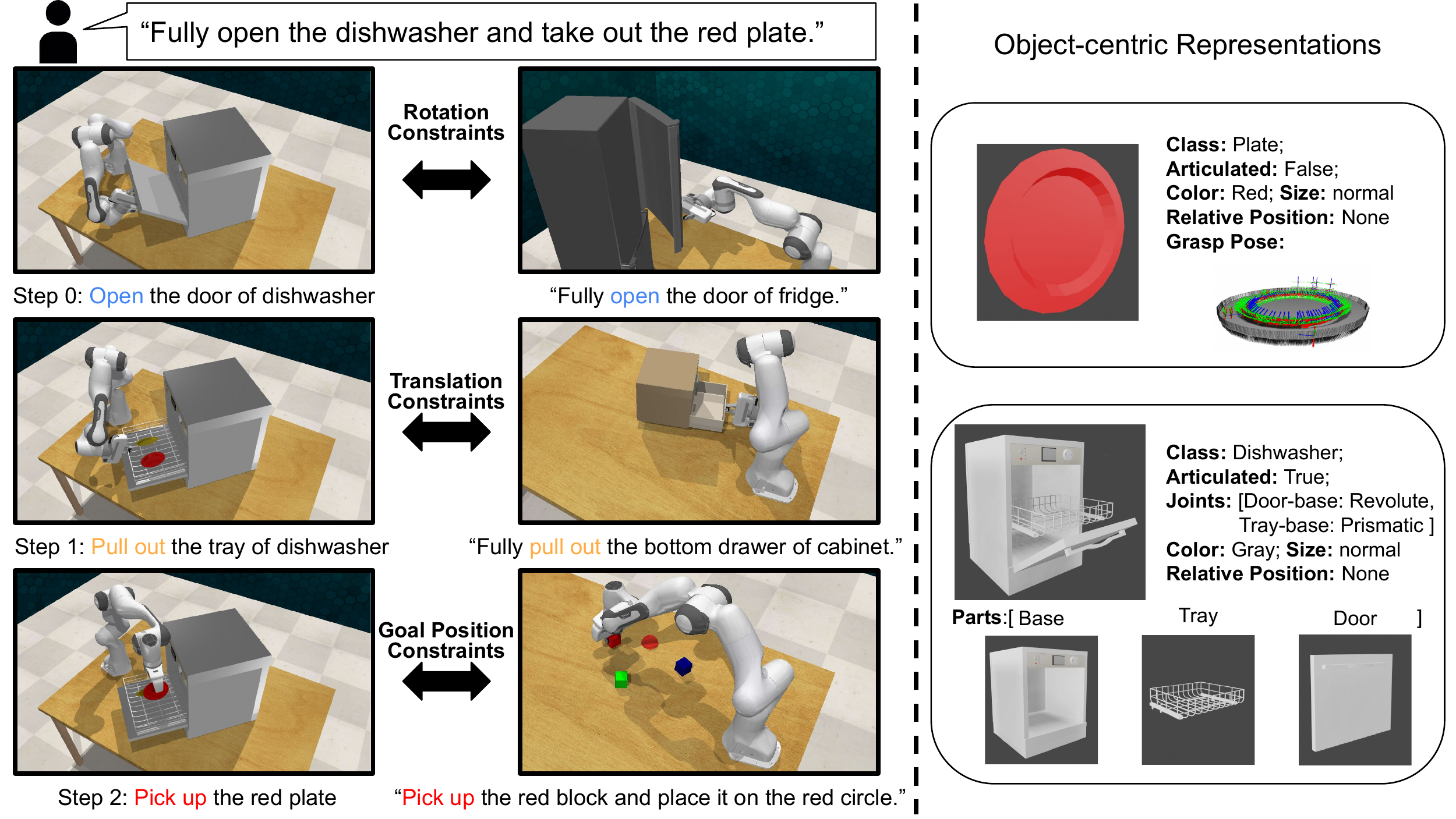}
\vspace{-4ex}
\caption{Given the language instructions and observations, the VLMbench requires the agent to generate an executable manipulation trajectory for specific task goals. On the left, we show that the complex tasks can be divided into the unit tasks according to the constraints of the end-effector, like ``Open the door of the dishwasher" and "Open the door of the fridge" should both follow the rotation constraints of the revolute joint. On the right, we show examples of object-centric representations, where all graspable objects or parts will generate local grasping poses as their attributes. Depending on the modular design, we can generate reasonable VLM data automatically. 
%\xin{where do you talk about this figure in the paper? It is unclear to me what this figure is used for.}
}
\label{fig:teaser}
\end{figure}

An ideal VLM benchmark should have at least three characteristics: The first one is scalability. Such a benchmark should automatically generate various physics-realistic 6 degrees of freedom (DoF) interactions with affordable objects and expand new tasks effortlessly. The second one is task categorization, which exploits commonality concerning robot motion between different semantic tasks and is almost ignored in existing works. The third one is reasonable language generation, which requires the benchmark can generate language instructions for testing diverse visual reasoning abilities without biases. However, existing benchmarks~\cite{szot2021habitat, james2020rlbench, yu2020meta, ahmed2020causalworld} lack at least one characteristic for VLM tasks. Motivated by these attributes, we present VLMbench, a highly categorical robotic manipulation benchmark with compositional language for visual reasoning. To build and scale VLMbench, we propose AMSolver, an automatic unit task builder that can compose unit tasks to create complex multi-step tasks and seamlessly adapt to novel objects. Compared to previous benchmarks, VLMbench categorizes manipulation tasks into various meta manipulation actions according to the constraints of robot trajectories for the first time. Meanwhile, the combinations of compositional language templates and object-centric representations provide numerous variations for visual reasoning in VLMbench, as shown in Figure~\ref{fig:teaser}.

To investigate the difficulty of the benchmark, we test them with several partially modal methods and a keypoint-based method, 6D-CLIPort, modified from the state-of-the-art language-guided manipulation method CLIPort~\cite{shridhar2022cliport}. The results show that there is still a massive room for improvement in the robust manipulation action generations and accurate language-guided visual understanding. 
To sum up, our contributions in this work include:
%\xin{too simple. You should highlight the novel contributions of this work.}
% \xin{(1) AMSolver, (2) VLM, (3) Model.}

\begin{itemize}
    \item AMSolver, an automatic demonstration generator for various task semantics, motion constraints, object types and states defined in a novel task template formulation.
    \item VLMbench, a robot manipulation benchmark on 3D tasks with visual observation and compositional language instructions, where we categorize the manipulation tasks by constraints and provide variations with minimal biases in the first time.
    \item 6D-CLIPort, a general vision-and-language manipulation baseline model evaluated on all kinds of VLMbench tasks.
\end{itemize}

\section{Related Work}
\label{sec:related}

\noindent\textbf{Robotic Manipulation Benchmarks~}
% \xin{This is related work section so you are supposed to discuss more about related works like the development of this field and what the existing benchmarks are. In the end, a few sentences may be used to specify the differences. Check out my VATEX paper for the related work section: https://arxiv.org/pdf/1904.03493.pdf.}
There are plenty of benchmarks proposed related to visual-language robotic tasks. ALFRED~\cite{ALFRED20} was proposed to do virtual object rearrangement tasks guided by visual observation and language instruction between different room-scale locations. ManipulaTHOR~\cite{ehsani2021manipulathor} introduced realistic interaction with objects using a 6 DoF configurable mobile manipulator. Habitat 2.0~\cite{szot2021habitat} and BEHAVIOR~\cite{srivastava2022behavior} incorporated real robot navigation with simple object interaction implementation for more comprehensive mobile manipulation tasks. CALVIN~\cite{mees2022calvin} collects 24 hours playing data with natural language instructions for long-horizon manipulation tasks. Regarding static manipulation benchmarks, MetaWorld~\cite{yu2020meta} collects a series of translation-only tasks with demonstrations for reinforcement learning. CausalWorld~\cite{ahmed2020causalworld} focused on causal inference within manipulation and showed examples on several simple object rearrangement tasks. RLBench~\cite{james2020rlbench} collected 100 different tasks by specifying robot arm end-effector waypoints for each of them for robot learning. Besides, there are also crowd-sourcing platforms that collect human demonstrations of a variety of common manipulation tasks through VR/AR devices, such as Robosuite~\cite{zhu2020robosuite} and Robomimic~\cite{mandlekar2021matters}. Compared to these works, our VLMbench includes both high-level task descriptions and low-level robot action translations and realizes automatic task builders for easier generation of complex tasks.

\begin{table}
\centering
\small
\resizebox{\textwidth}{!}{%
\begin{tabular}{@{}lccccc@{}}
\toprule
\textbf{Benchmark} &
  \begin{tabular}[c]{@{}c@{}}\textbf{novel object}\\ \textbf{adaptation}\end{tabular} &
  \begin{tabular}[c]{@{}c@{}}\textbf{automatic} \\\textbf{trajectory} \textbf{generation}\end{tabular} &
  \begin{tabular}[c]{@{}c@{}}\textbf{variant}\\ \textbf{object} \textbf{property}\end{tabular} &
  \begin{tabular}[c]{@{}c@{}}\textbf{automatic} \\\textbf{6-DoF} \textbf{grasping}\end{tabular} &
  \begin{tabular}[c]{@{}c@{}}\textbf{constraint-based}\\\textbf{task} \textbf{formulation}\end{tabular} \\
  \midrule
CausalWorld~\cite{ahmed2020causalworld}  & \xmark & \xmark & \cmark & \xmark & \xmark \\
MetaWorld~\cite{yu2020meta}     & \xmark & \cmark & \xmark & \xmark & \xmark \\
ManipulaTHOR~\cite{ehsani2021manipulathor}       & \xmark & \xmark & \cmark & \xmark & \xmark \\
%ALFRED~\cite{ALFRED20}       & \cmark & \xmark & \xmark & \xmark & \xmark & \xmark \\
Habitat 2.0~\cite{szot2021habitat}   & \cmark & \cmark & \cmark & \xmark & \xmark \\
Robomimic~\cite{mandlekar2021matters}    & \xmark & \xmark & \xmark & \xmark & \xmark \\
BEHAVIOR~\cite{srivastava2022behavior}      & \xmark & \cmark & \cmark & \xmark & \xmark \\
RLBench~\cite{james2020rlbench}      & \xmark & \cmark & \cmark & \xmark & \xmark \\
CALVIN~\cite{mees2022calvin}    & \xmark & \xmark & \xmark & \xmark & \xmark \\
\midrule
VLMbench (ours)  & \cmark & \cmark & \cmark & \cmark & \cmark \\ 
\bottomrule
\end{tabular}%
}
\label{tab:benchmark}
\caption{Comparison of existing robotic manipulation benchmarks to VLMbench.}
\vspace{-0.5cm}
\end{table}

\noindent\textbf{Vision and Language Tasks for Embodied AI~}
Tasks such as Vision Question Answering (VQA)~\cite{antol2015vqa,das2018embodied}, Video Captioning~\cite{gao2017video,wang2018reconstruction}, Image-Text Retrieval~\cite{zhang2020context,chen2020imram}, etc., connect vision and NLP research to produce semantics from combined embedding features. Recent works such as referring object spatial relations~\cite{liu2019clevr}, dynamic events~\cite{yi2019clevrer} provide valuable reasoning result for the agent to make actions. Vision-and-Language Navigation (VLN)~\cite{anderson2018vision,lu2019vilbert,pashevich2021episodic,kurita2020generative} proceeds another step to use visual observations and language instructions for robotic navigation directly.
Recent benchmarks~\cite{ALFRED20,szot2021habitat,srivastava2022behavior} combined VLN with abstracted object interaction to include more object rearrangement tasks. Compared to the tasks mentioned above, our benchmark is designed for more complex robotic manipulation that simulates physics-realistic 6DoF grasping and requires the reasoning ability from the visual-language space to the executable action space. %modular unit tasks that are generalizable across similar examples and can compose to variate compositions.
% \xin{Unclear. What is static robotic manipulation? I do not think any VLN tasks do manipulation at all (some of them do interact with objects though). You should explain the differences.}

\noindent\textbf{Language-Instructed Manipulation~}
Recently, various manipulation tasks have been researched with language input either describing the entire task, or serving interactive input for task specifications. HULC~\cite{mees2022matters} proposes a hierarchical network for long-horizon manipulation tasks, which contains multi-modal transformers for task planning and a policy network for action generation.
Structformer~\cite{liu2021structformer} proposes an object selection network from language and visual encodings, as well as a language conditioned pose generator for semantic object rearrangement. Stepputtis et al.~\cite{stepputtis2020languageconditioned} proposed a closed-loop control model for pouring tasks. CLIPort~\cite{shridhar2022cliport} proposed a two-stream framework to learn a spatial attention map for 2D object manipulations.
Lynch et al.~\cite{lynch2020language} fed natural language instructions to a goal-conditioned policy pretrained using imitation learning for various tasks.
INVIGORATE~\cite{zhang2021invigorate} proposed an interactive system that takes language input to correct false estimations and re-plan for object grasping in clutter.
Shao et al.~\cite{shao2021concept2robot} trained a joint language-vision model on 7-DoF goal trajectory estimation with another task classifier for multi-task training. Goyal et al.~\cite{goyal2021zero} learned zero-shot task adaptation to accomplish novel tasks through differences described in natural language. In this paper, our baseline evaluated on VLM is built over CLIPort, which learns full 6D grasping included, variate-length robot manipulation tasks from language instructions.

\section{AMSolver: Automatic Manipulation Solver}
\label{sec:amsolver}
\begin{figure}[!tbp]
% \flushright
\centering
\includegraphics[width=\columnwidth]{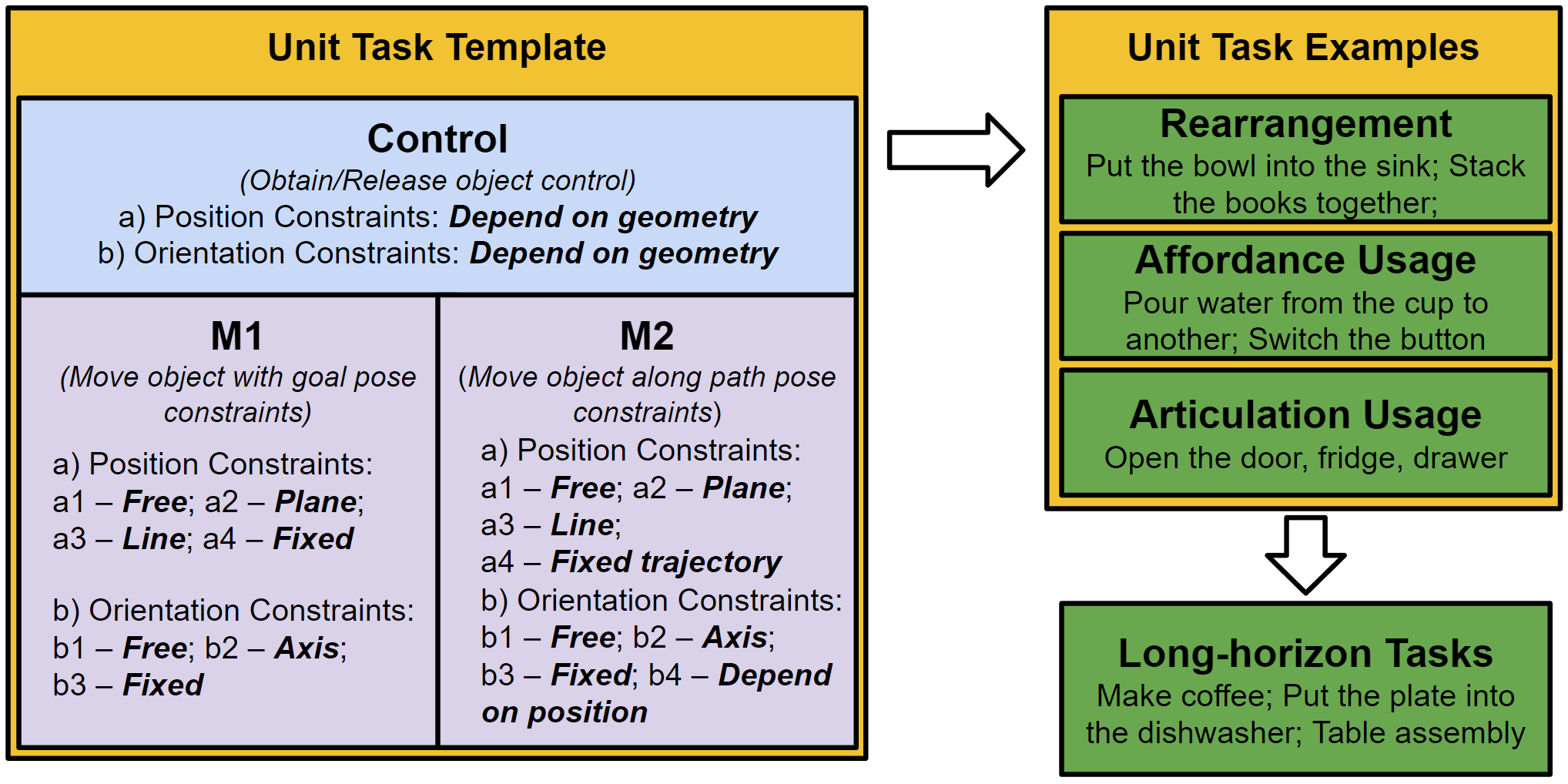}
\vspace{-4ex}
\caption{The unit task templates of AMSolver. On the left, we show three unit task templates parameterized by position and orientation constraints over the robot end-effector on either goal pose or entire path. By combining these unit task templates, various task examples can be generated. For example, on the right we show three main common task types of household tasks and long-horizon tasks composed of unit tasks. }
% \vspace{-4ex}
\label{fig:unit_task}
\end{figure}

We consider a rule-based task categorization that supports most of daily manipulation tasks, which follows a unified formulation. To this extent, we propose Automatic Manipulation Solver (AMSolver)\footnote{AMSolver is implemented in CoppeliaSim~\cite{6696520} (Free Educational License) and codes are based on RLbench~\cite{james2020rlbench} (RLBench Software License) and PyRep~\cite{james2019pyrep} (MIT License).}. We focus on simple unit tasks and introduce a unit task template that categorizes motion constraints and generates a wide range of household tasks (see examples in Figure~\ref{fig:unit_task}). The formulation treats objects as a representation that could have variations in appearance, and enables automatic demonstration generation using off-the-shelf task and motion planners.

% We consider three main categories of household manipulation tasks: rearrangement (move an object from place to place, e.g., collect trash into the bin), affordance usage (exploit particular usage of some objects, e.g., pour water from a mug), and multi-step tasks (can be decomposed into several individual steps, e.g., connect every two pieces to assemble a table from parts). To formulate a unified task representation, we assume that every task within these categories can be considered as a combination of objects and unit actions. To describe the essential elements of tasks, we proposed Automatic Manipulation Solver (AMSolver)\footnote{AMSolver is implemented in CoppeliaSim~\cite{6696520} (Free Educational License) and codes are based on RLbench~\cite{james2020rlbench} (RLBench Software License) and PyRep~\cite{james2019pyrep} (MIT License).}, a rule-based automatic task builder consisting of object-centric representations and unit task templates, as shown in Fig.~\ref{fig:unit_task}. Below we illustrate how a complex task can be built bottom-up.
% We explain how complex tasks can be built based on the proposed concepts in the following paragraphs.

\subsection{Rule-based Unit Tasks}
Since the tasks have notable variations, we assume that each complex task can be decomposed into the combinations of several unit tasks from the aspects of end-effector trajectories. We define a \textit{unit task} as the semantic step of completing a sub-goal of the entire task. 
% A \textit{unit task} should keep an invariant rule among different tasks, consisting of a simple structure, and have enough scalability to represent every task step. 
Specifically, a unit task is defined in a formula of `take action on an object under certain constraints' where a unit task can be parameterized by two constraints: (a) \textbf{position constraints} and (b) \textbf{orientation constraints}, which describe the valid spatial space or orientation range, respectively, of the end-effector for a specific task.
We propose three unit task templates detailed below that can compose the aforementioned complex tasks.
\\
\noindent\textbf{(1) Control} is a preparation or ending step of a task, which models the transition of the object state, where the state indicates whether the robot can move the object or not. In this work, we specify one way of transition: to obtain control by grasping it and release control by opening the gripper. There are other ways to obtain control, like pushing, hitting, etc. However, these transitions will naturally lead to the constraints in the following sub-tasks, so they cannot be considered a general component for any complex task.
% Due to the generalization, we only consider grasping in this work. 
In this unit task, the position and orientation constraints depend on the geometry of object instances. 
% a robot either can \textbf{a1} grasp objects, or get in contact with objects to begin to \textbf{a2} push; to release control, the robot can \textbf{a3} release objects from grasping or detach contact with objects from pushing. For these two basic actions, we assume every object used in the task is \textit{pushable} (so it's not mentioned in object property), while only some objects are \textit{graspable}. Examples of this unit task include \textbf{T0(a1,b1)}: grasping a graspable object, and \textbf{T0(a2,b2)}: beginning to push an object that is not graspable.
\\
\noindent\textbf{(2) M1} denotes moving the target object with goal pose constraints, which can be modeled as a 6 DoF transform in the robot's workspace. The position constraints define a bounded goal space in $\mathbb{R}^3$, while the orientation constraints define a valid 3D orientation $SO(3)$. We consider four types of position constraints: (a1) \textit{Free}, (a2) \textit{Plane}, (a3) \textit{Line}, and (a4) \textit{Fixed}, which means the goal position is any point inside a 3D space (a1), constrained in a 2D plane (a2), constrained in a line-shape area (a3), or fixed to a certain point (a4). There are three types of orientation constraints: (b1) \textit{Free}, (b2) \textit{Axis}, and (b3) \textit{Fixed}, which means the goal orientation is unlimited in 3D rotation space (b1), only rotated along an axis in space (b2), or fixed at a given orientation (b3). For example, M1[a2, b1] can represent placing the object on a tabletop (with plane position constraint), while M1[a3, b2] can represent moving the object to a position on one line and ending with a constrained orientation of one axis, like dropping a stick into the hole.
\\
\noindent\textbf{(3) M2} denotes moving the target object along a trajectory while satisfying the motion constraints during the entire path, which implies a more strict condition than \textbf{M1}. The constraints are mostly from object articulation, like revolute joints on doors, or task-specific requirements, like keeping the opening upward for a full filled mug. Therefore, there are four kinds of position constraints: (a1) \textit{Free}, (a2) \textit{Plane}, (a3) \textit{Line}, and (a4) \textit{Fixed trajectory}, which means any position inside a space (a1), a plane (a2), a line (a3) or a predefined path (a4) should be feasible for the trajectory, and four kinds of orientation constraints: (b1) \textit{Free}, (b2) \textit{Axis}, (b3) \textit{Fixed}, and (b4) \textit{Depend on position}, which means every pose in the trajectory should be unlimited orientations (b1), at most rotated along an axis (b2), fixed to a certain orientation (b3), or depended on the corresponding positions. For example, M2[a2, b2] means moving the object inside a 2D plane while maintaining the orientation of one axis, like wiping the table, while M2[a4, b2] means moving the object along a fixed trajectory with maintaining the orientation of one axis, like using a screwdriver to tighten a screw. It is worth mentioning that M2[a1,b1] will degenerate to M1[a1,b1]. According to our unit task definition, we have covered every feasible 6 DoF poses of the end-effector. Therefore, we have reason to believe that these unit tasks can represent any complex task in the action space.

\subsection{Object-centric Representation}
Some recent works~\cite{li2021igibson,yuan2022sornet} have used object-centric representations for manipulation. Since the properties are defined on objects, these representations can easily cross the variations of environments, agents, and tasks. Our benchmark assumes that objects used in the tasks are rigid, and their fundamental properties will not change during the tasks. Therefore, we can parameterize the objects as a set of configurations, including class, color, size, and geometry shape. If the object is articulated, its whole configuration will contain the configuration of each part and the physical constraint of each connection. For example, a door consists of three parts: the door base, plank, and handle, so its configuration will contain each part's configuration and record the positions and ranges of two revolute joints.

\subsection{Automatic Demonstration Generation}

To create a task example, the user could compose a task template from the unit library of formulation, such as a unit task of object rearrangement as \textbf{Control}-\textbf{M1} or a more complex task of object stacking as \textbf{Control}-\textbf{M1}-\textbf{Control}-\textbf{M1}-\textbf{Control}-\textbf{M1}, etc., and then select the objects from a given set to be included in the scene as object to be manipulated and distractor objects. In the simulation the object placement could be randomized with custom specification. The corresponding language descriptions could also be generated from templates.

To implement \textbf{Control} as grasping, we create an object-wise grasp pose dictionary. Specifically, a point cloud-based grasping pose detection algorithm~\cite{ten2017grasp} is implemented to search all feasible grasping poses, given the object's shape and robot gripper parameters. The grasp poses are saved and transformed to world space for a particular task by simply multiplying with the object's pose. To implement \textbf{M1} and \textbf{M2}, we integrate customized motion constraints in the OMPL motion planner library so that the calculated trajectory satisfies the constraints automatically. For more details and task examples, please refer to Appendix \ref{app:impl}.

\section{VLMbench: Visual-and-Language Manipulation Benchmark}
\label{sec:vlmbench}
% In this section, we will introduce VLMbench in detail. Instead of handcrafting many tasks, we classify tasks into different task categories by constraints. Also, we will describe the hierarchical language instructions in this benchmark.

\begin{figure}[tbp]
\centering
\includegraphics[width=0.9\columnwidth]{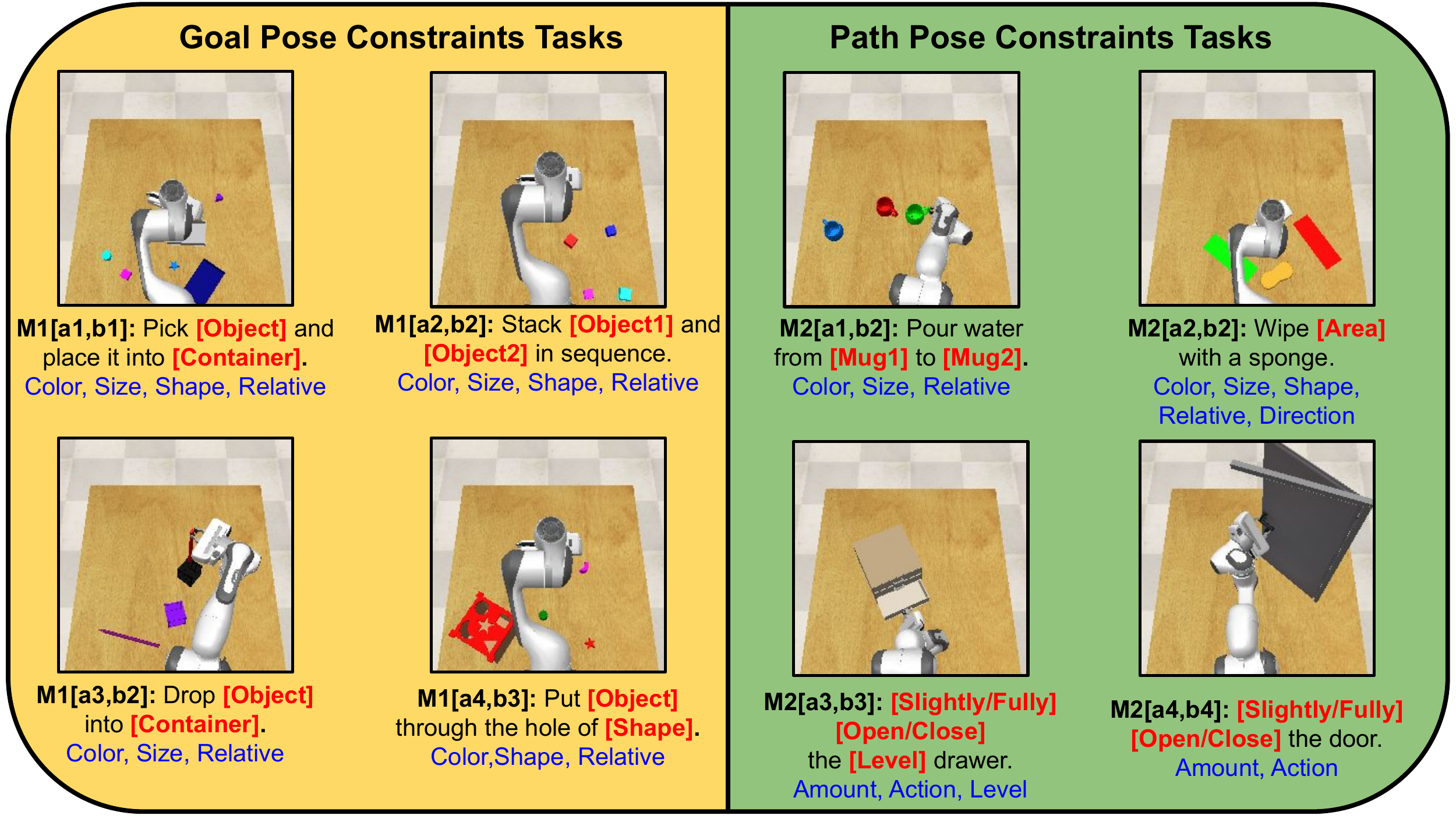}
\vspace{-2ex}
\caption{We show four task categories with goal-pose constraints on the left side, and on the right side, we show four task categories with path-pose constraints. For each task category, we visualize the observations of the overhead view and list the main unit task, variations, and instruction templates. The red words with brackets indicate the blank where the variations descriptions can fill in. The blue words indicate the variations in the tasks. The combinations of variations will lead to various instance-level tasks.}
% \vspace{-4ex}
\label{fig:tasks}
\end{figure}

\subsection{Problem Definition}
Given language instructions, the Vision-and-Language Manipulation (VLM) task requires an embodied agent to follow the instructions to complete tabletop manipulation tasks. Formally, at the beginning of the task, the agent receives a set of language instructions $L=\{L_1, L_2, ..., L_n\}$, where $L_i$ denotes one sentence of arbitrary length. The initial state $s_0$ contains multi-view RGB images, depth images, segmentation information, and robot states, including joint angles, velocities and torques, and end-effector pose. Given the observations and language instructions, the agent needs to estimate an executable action command $a_0$, directly working on the end-effector or joints. Then, at each step $t$, the agent receives new observations $o_t$ and generates the action $a_t = f(s_t|s_0, s_1, ..., s_{t-1},L)$ for the next step. The step loop will repeat until the agent sends a stop action or should be terminated, e.g., achieve the success conditions or the limitation steps. The agent should obey the constraints provided by language instructions during the whole running.

\subsection{Tasks and Dataset}

In previous works, the researchers manually designed manipulation tasks by implicit prior knowledge without categories. Instead, we are trying to build tasks from the perspectives of elementary manipulation abilities. In other words, since different tasks have various semantic meanings, we consider the task with the same unit tasks combination should be in the same category from the aspect of the action space. For example, ``Open the door of the fridge" and ``Open the door of the microwave" require the same action ability except for semantic meanings and grasping poses which depend on the object geometry. Therefore, we define eight general task categories, represented by one typical task in each category, shown in Table~\ref{table:task_build} and Fig.~\ref{fig:tasks}. We use the definitions in the unit task templates to represent the main constraints of each task category. The task details can be found in Appendix~\ref{app:task} and dataset statistics can be found in Appendix~\ref{app:dataset}.

\noindent\textbf{Task Variations}
The manipulated object's properties can randomly change for each task category, and every combination leads to a task instance. In the VLMbench, we use 8 different variations: color, size, relative position, shape, direction, level, amount, and action type. The variations are from two perspectives: object and motion. Object variations include color, size, shape, relative position, and direction. Here, the \textbf{color} is chosen from 20 colors in the seen settings. The \textbf{size} contains the relative descriptions between two objects, ``smaller" and ``larger", and descriptions between three objects, which are ``large",``medium", and ``small". The \textbf{shape} contains 5 types of objects for the seen and unseen settings. The \textbf{relative position} describes the spatial relationship between two objects, like the top, front, rear, left, and right. The \textbf{direction} contains two descriptions for rectangular prism in a plane:``horizontal" and ``vertical". For the objects that have the vertical structure, the \textbf{level} includes ``top", ``middle," and ``bottom." From the motion view, the variations are amount and action type. The \textbf{amount} means how far the task needs to be done, consisting of ``fully" and ``slightly." The \textbf{action type} includes ``open" and ``close", especially for the articulated objects. The table of these variations and models used can be found in Appendix~\ref{app:task}.

\noindent\textbf{Unseen Settings} 
All tasks in the unseen settings are unseen <color,shape> combinations from an unseen color collection and an unseen shape collection (where the shapes include all object classes and variants). The unseen color collection has five new colors that do not appear during training, including brown, gold, pink, chocolate, and coral. As for the unseen shape collection, it has some overlap with the seen library for the tasks with color variations (but the <color,shape> combinations are always unseen), and is exclusive for the tasks without color variations (e.g., we introduce a new door model with a rotatable handle for the unseen setting of Door tasks). So there are mainly three kinds of unseen combinations: <unseen color,unseen shape>, <unseen shape>, and <unseen color,seen shape>. The exact object models used for each task can be found in Appendix~\ref{app:task}.

\begin{table}[!tbp]
\centering
\label{table:task_build}
\resizebox{\textwidth}{!}{%
\begin{tabular}{cccc}
\toprule
\textbf{Task Categories} &
  \textbf{Main Constraints} &
  \textbf{Variations} &
  \textbf{Instructions Samples} \\ \midrule
Pick \& Place objects &
  \begin{tabular}[c]{@{}c@{}}M1 -- Position: Free,\\  Orientation: Free\end{tabular} &
  \begin{tabular}[c]{@{}c@{}}Color, Size, Shape, \\ Relative Position\end{tabular} &
  \begin{tabular}[c]{@{}c@{}}``Pick the red cube and place\\  it into the green container."\end{tabular} \\ \midrule
Stack objects &
  \begin{tabular}[c]{@{}c@{}}M1 -- Position: Plane, \\ Orientation: Axis\end{tabular} &
  \begin{tabular}[c]{@{}c@{}}Color, Size, Shape,\\  Relative Position\end{tabular} &
  \begin{tabular}[c]{@{}c@{}}``Stack the small star and \\ the medium star in sequence."\end{tabular} \\ \midrule
Drop pencil &
  \begin{tabular}[c]{@{}c@{}}M1 -- Position: Line,\\  Orientation: Axis\end{tabular} &
  Color, Size, Relative Position &
  \begin{tabular}[c]{@{}c@{}}``Drop the left pencil \\ into the right container."\end{tabular} \\ \midrule
Put into shape sorter &
  \begin{tabular}[c]{@{}c@{}}M1 -- Position: Fixed, \\ Orientation: Fixed\end{tabular} &
  Color, Shape, Relative Position &
  \begin{tabular}[c]{@{}c@{}}``Put the triangular prism \\ through the hole of triangle."\end{tabular} \\ \midrule
Pour water &
  \begin{tabular}[c]{@{}c@{}}M2 -- Position: Free, \\ Orientation: Axis\end{tabular} &
  Color, Size, Relative Position &
  \begin{tabular}[c]{@{}c@{}}``Pour the water from \\ the green mug to the red mug."\end{tabular} \\ \midrule
Wipe table &
  \begin{tabular}[c]{@{}c@{}}M2 -- Position: Plane,\\  Orientation: Axis\end{tabular} &
  \begin{tabular}[c]{@{}c@{}}Color, Size, Directions, \\ Relative Position\end{tabular} &
  \begin{tabular}[c]{@{}c@{}}``Wipe the horizontial area \\ with a sponge."\end{tabular} \\ \midrule
Use drawer &
  \begin{tabular}[c]{@{}c@{}}M2 -- Position: Line, \\ Orientation: Depend on position\end{tabular} &
  Amount, Action Type, Level &
  ``Fully close the top drawer." \\ \midrule
Use door &
  \begin{tabular}[c]{@{}c@{}}M2 -- Position: Fixed trajectory, \\ Orientation: Depend on position\end{tabular} &
  Amount, Action Type &
  ``Slightly open the door." \\ \bottomrule
\end{tabular}%

}
\caption{The table contains the category-level tasks in our dataset, with their main constraints, variations and instruction samples.}
% \vspace{-14px}
\end{table}

\section{Vision-and-Language Manipulation Agent}
\label{sec:agent}
\begin{figure}[tbp]
% \flushright
\centering
\includegraphics[width=\columnwidth]{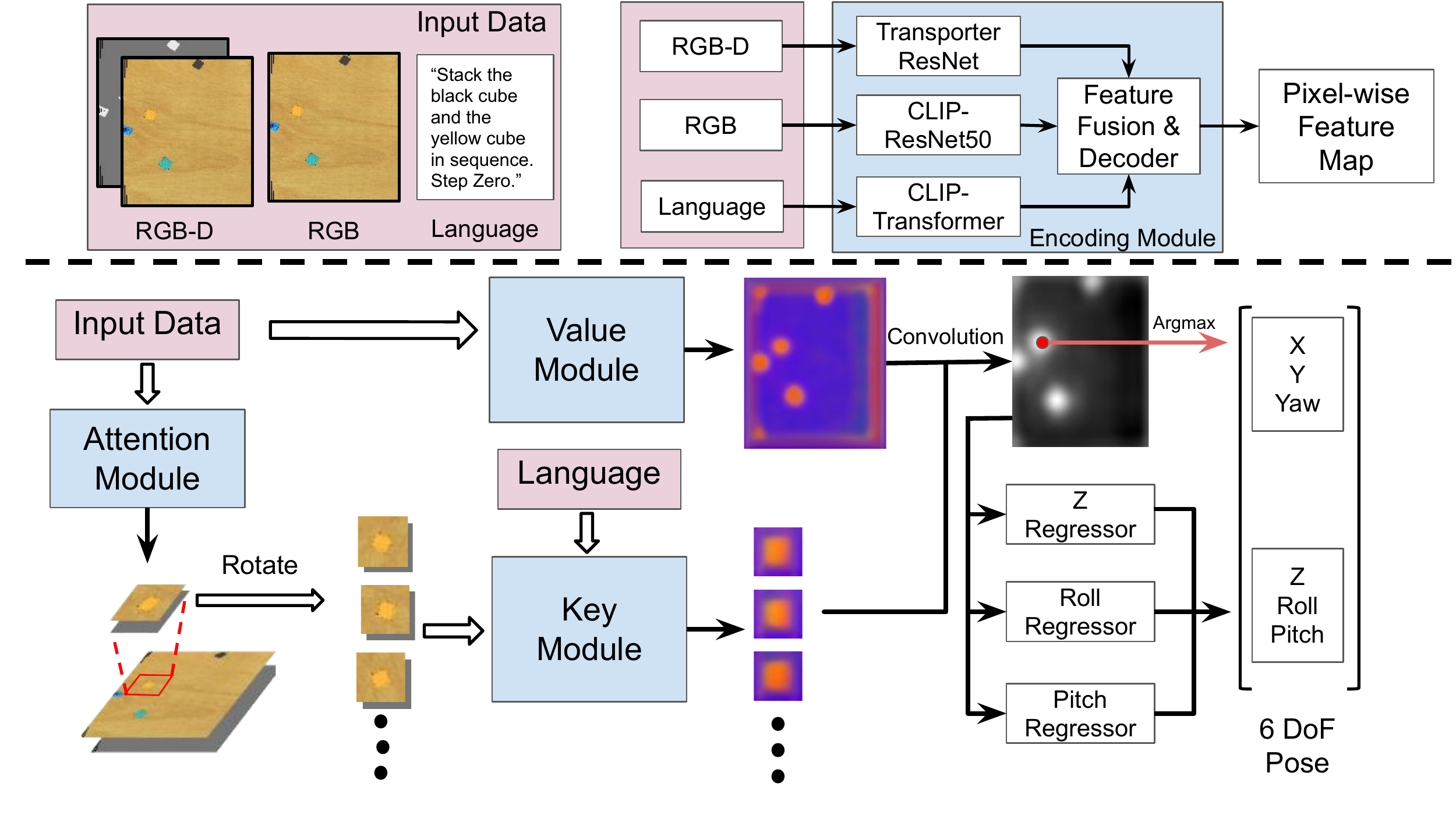}
\vspace{-4ex}
\caption{The structure of 6D-CLIPort. Please refer to Sec.~\ref{sec:agent} for details.}
% \vspace{-4ex}
\label{fig:model}
\end{figure}

To provide a baseline method that can solve VLM tasks, we propose a neural-network based agent 6D-CLIPort that takes in the input of multi-view RGB-D observations and task language descriptions and outputs 6 DoF pose keypoints along the path that can accomplish a specific task. For instance, for pick-and-place task, the agent will iterate twice and output the object's 6D pose for pick, and place, separately.

% \subsection{6D-CLIPort}\label{6dcliport}

The overall flow of 6D-CLIPort\footnote{6D-CLIPort is implemented based on CLIPort~\cite{shridhar2022cliport} (Apache-2.0 License)} is shown in Figure~\ref{fig:model}. The input data passes through visual and language encoders and the embedded features are fused together to obtain a pixel-wise feature map that shows the probability of object of manipulation interest. The encoding module is reused in 3 models (Attention, Value, and Key Module) as follows: The Attention module takes maximum of the feature map to get 2D image sample crops. The Key module takes input of the rotated crops and output feature map crops. Then, the feature map crops are used as convolution kernels and pass through the feature map of the entire image from Value module to get a 3D pose heatmap (x-y 2D position and yaw rotation, and their estimations are obtained by maximizing the probability). Three fully connected neural network modules will regress one of the remaining 3 dimensions each (z position and roll, pitch rotation) from this heatmap. Finally, a full 6 DoF pose is composed. Compared to the state-of-the-art method CLIPort~\cite{shridhar2022cliport}, we resolved its two constraints: 1. increase 3 DoF only motions to full 6 DoF; 2. output arbitrary number of poses than fixed 2 outputs for pick and place actions. Besides, we introduce some details below.

\noindent\textbf{Multi-view Vision Fusion}
To get a better RGB-D input, we first fuse RGB-D input from several cameras with known poses into a 3D colored point cloud, and then project it along the vertical direction facing towards the table plane to get a top-view RGB-D image.
% In order to utilize the visual information of the cameras from different views, the model turns the RGB-D images to the point cloud with colors, crops the points above the table, and projects the cropped points to an overlooking view. Then, we can get canonical color maps and depths.

% \noindent\textbf{Feature Extraction}
\noindent\textbf{Encoding Module}
 The agent has two feature extraction streams: semantic and spatial streams. The agent uses the pretrained CLIP's ResNet50 and Transformer model~\cite{radford2021learning} to encode the RGB and languages in the semantic stream. An untrained Transporter ResNet~\cite{zeng2020transporter}, which has 43 layers and 8 strides, is used in the spatial stream to encode the RGB-D image. Then, the decoder fuses these latent space features by concatenation, fully convolution, and up-sampling layers, and predicts a dense pixel-wise feature map. More details can be found in~\cite{shridhar2022cliport}.

% \noindent\textbf{Pose Prediction with Attention Mechanism}
% There are three modules with the same structure in the agent: Attention Module, Value Module, and Key Module. The Attention Module estimates an attention map $A:\mathbb{R}^{H\times W}$ that has a high probability of the position of the manipulated object in the image coordinate. Then the original RGB-D image will be cropped by a fixed size, rotated with the predefined angles, and fed into the Key module with language instructions, where the outputs are the key feature maps. Meanwhile, the original input data will pass through the Value Module to obtain the value feature map. By convolution, the agent can get the pose heatmap $Y:\mathbb{R}^{H\times W\times k}$ of high probability 2D position and the rotation along the perpendicular axis to the image plane with $k$ discrete rotations. To gain other degrees of freedom, three 3-layers FCNN regressors input the heatmap cropped by the predefined range of high probability 2D position and the rotation and output the z values and other two rotations. All rotations in the agent have been discretized into $k$ portions. In this work, the $k=36$. Except the z regressor use the Huber loss function~\cite{huber1992robust}, other losses for the attention map $A$, the pose heatmap $Y$, and two rotation regressors are using the cross-entropy with softmax~\cite{shridhar2022cliport}.

% \subsection{Training Details}
\noindent\textbf{Implementation Details}
% \subsection{Implementation Details}
We separately train the agent for each task category. For example, the agent for pick and place tasks will jointly train on the data of all variations mentioned in Table~\ref{table:task_build}. 
In details, since the VLMbench is built by the unit task templates, the input demonstrations $D = \{L, \zeta_1, \zeta_2, ... ,\zeta_i\}$ can be divided into different sub demonstrations by the waypoints generated from unit task templates, where each step $\zeta_i=(L_i, o_i, g_i)$ consists of language instruction $L_i$, observation $o_i$, and the 6 DoF sub-goal waypoint pose $g_i$ for the current step. For step $i$, we use the observations in the first frame of this step as $o_i$ and prompt high-level instruction $L$ with ``Step i" as $L_i$. The input RGB-D image has resolution $160\times128$. The feature heatmap is with dimension $16$ in \textit{x-y} 2D plane, and the crops from Attention Module are rotated 36 times before feeding into the Key Module.

\section{Experiments}
\label{sec:experiments}

\subsection{Experimental Setup}
\label{sec:experimental_setup}
\noindent\textbf{Evaluation Settings~}
Before testing each baseline, we preprocess the tasks to help the agent eliminate trivial steps. 1) we divided tasks into the sub-goal sequences by the ground truth waypoints generated by AMSolver, so that the agent only needs to estimate the actions of the predefined unit task sequence for each task. 2) Since we have the ground truth gripper state for each sub-goal, we also provide whether the gripper should open or close for each action estimation. 3) To increase the grasping stability, we use the pre-generated grasping pose instead of estimations if these two poses are close enough (distance is less than 5 cm and the rotation is less than 10 degrees). 4) To reduce the failure grasping cases due to the motion planning, we use a predefined pre-grasping offset, which is 8cm backward along the z-axis of the target grasping pose, and a post-grasping offset, which is 8cm upward along the z-axis of the world frame.

In each episode of one task variation, the simulator imports a test configuration from the pre-collected test dataset, which includes the initialization poses of objects, the success conditions of the task, and a set of waypoints that can finish the task for reference.
The agent should solve the task in the online simulator within a limited number of steps.
Success rate is used as the primary evaluation metric, calculated by dividing the number of success conditions satisfied by the number of tests. We use the average success rate of all variations for each task category. The success conditions are mainly determined by an object detector or a joint detector. The object detector returns true if particular objects have moved inside the predefined space, and the joint detector returns true when the joint angle reaches the predefined range. The success conditions of each task can be found in Appendix~\ref{app:task}.

\noindent\textbf{Baselines~}
In addition to the 6D-CLIPort model, we provide two kinds of baseline models for comparison, one with partial input modalities and the other with partial ground-truth predictions.

To test the influence of different input modalities, we traine two other agents with partial modalities: a \textit{Language-Only} agent and a \textit{Vision-Only} agent. These agents use the same model architecture as 6D-CLIPort but have different input modalities. The Language-Only agent uses the CLIP transformer for language encoding and the visual inputs are all zeros. The Vision-Only agent use the same RGBD inputs as in the 6D-CLIPort agent, but its language input will only include the prompt for steps indication like ``Step zero'' without any high-level language instructions.

To measure the capabilities and limitations of 6D-CLIPort, we individually test its position or orientation estimation abilities by giving the ground-truth values of the other. Although the trajectories of finishing the tasks are various and we cannot obtain the optimal position and rotation information, we can regard the poses of waypoints as sub-optimal solutions. \textit{GT Pos} means given ground truth \textit{x/y/z} positions while the other three parameters for 3D orientation are estimated, and \textit{GT Ori} suggests the contrary (known orientation, using estimated 3D position).

\begin{table}[tbp]
    \centering
    \begin{adjustbox}{max width=\textwidth, center}{
    \begin{tabular}{lrrrrrrrr}
    \toprule
         & \multicolumn{2}{c}{\textbf{Pick\&Place}} & \multicolumn{2}{c}{\textbf{Stack}}
         & \multicolumn{2}{c}{\textbf{Drop}} & \multicolumn{2}{c}{\textbf{Shape Sorter}}
         \\
         \cmidrule(lr){2-3} \cmidrule(lr){4-5} \cmidrule(lr){6-7} \cmidrule(lr){8-9} 
         Agent & Seen & Unseen
                & Seen & Unseen
                & Seen & Unseen
                & Seen & Unseen \\
        \toprule
        Language-Only
        & 0.00 & 0.00
        & 0.00 & 0.00 
        & 0.00 & 0.00 
        & 0.00 & 0.00
        \\
        Vision-Only 
        & 6.31 & 9.85
        & 6.89 & 1.79
        & 0.00 & 0.00
        & 0.00 & 0.33
        \\
        6D-CLIPort 
        & \textbf{28.28} & \textbf{27.53} 
        & \textbf{22.19} & \textbf{18.37} 
        & \textbf{6.42} & \textbf{6.42}
        & \textbf{17.33} & \textbf{12.33}
        \\
        \midrule
        6D-CLIPort (GT Ori)
        & 28.03 & 26.26
        & 26.53 & 26.02
        & 17.91 & 16.22
        & 24.00 & 15.67 \\
        6D-CLIPort (GT Pos)
        & 83.84 & 75.25
        & 58.93 & 50.51
        & 16.89& 11.82
        & 18.00 & 17.33 \\
         \bottomrule
         & \multicolumn{2}{c}{\textbf{Pour}} & \multicolumn{2}{c}{\textbf{Wipe}}
        &\multicolumn{2}{c}{\textbf{Door}} & \multicolumn{2}{c}{\textbf{Drawer}} \\
        \cmidrule(lr){2-3} \cmidrule(lr){4-5} \cmidrule(lr){6-7} \cmidrule(lr){8-9} 
          & Seen & Unseen
                & Seen & Unseen
                & Seen & Unseen
                & Seen & Unseen \\
        \toprule
        Language-Only
        & 0.00 & 0.00
        & 0.00 & 0.00
        & 0.00 & 0.00
        & 4.17 & 1.04
        \\
        Vision-Only 
        & 0.00 & 0.00
        & 19.80 & 20.80
        & 0.00 & 0.00
        & 14.58 & 7.29
        \\
        6D-CLIPort 
        & \textbf{1.00} & \textbf{1.00}
        & \textbf{22.40} & \textbf{21.00}
        & \textbf{6.00} & \textbf{5.00}
        & \textbf{22.92} & \textbf{15.63}
        \\
        \midrule
        6D-CLIPort (GT Ori)
        & 3.67 & 3.67
        & 25.80 & 25.20
        & 6.00 & 5.00
        & 23.96 & 17.71 \\
        6D-CLIPort (GT Pos)
        & 0.33 & 0.67
        & 60.20 & 53.40
        & 27.00 & 27.00
        & 43.75 & 52.08 \\
         \bottomrule
    \end{tabular}}
    \end{adjustbox}
    \caption{Success rates of all task categories, including both seen and unseen settings.}
    \label{tab:results_main}
    % \vspace{-0.5cm}
\end{table}

\subsection{Result Analysis}
\noindent\textbf{Main Results on Different Task Categories and Variations~}
The main results on different task categories are shown in Table~\ref{tab:results_main}. 
We observe that 6D-CLIPort performs better on the tasks that have lower rotation variances, including ``Pick\&Place," ``Stack,"``Shape Sorter," ``Wipe," and "Drawer". It indicates that 6D-CLIPort can better estimate the positions than orientations in the 3D spaces. Moreover, 6D-CLIPort performs poorly on the ``Pour" tasks since the task needs to adjust the pouring poses following the grasping pose, which introduces additional difficulties.
Moreover, although the success rates in the unseen settings are generally lower than those in the seen settings, the performance drop is reasonable and not dramatic, showing that 6D-CLIPort can transfer the learned manipulation knowledge from seen objects to unseen objects, benefiting from the powerful transfer ability of the pre-trained CLIP model~\cite{radford2021learning}.

We also show the results from the perspective of variations across task categories in Table~\ref{tab:results_main_sub}.
The results show that the agent is more sensitive to the novel compositions and thus has larger seen-unseen performance drop on variations such as ``Shape'', ``Level'', ``Action Type'', and ``Amount''.

\noindent \textbf{Impact of Different Input Modalities~}
Table~\ref{tab:results_main} also compares 6D-CLIPort with Language-Only and Vision-Only agents, demonstrating the impact of different input modalities.
(1) The baseline vision-and-language manipulation model 6D-CLIPort performs the best on all tasks, showing the importance of both visual observations and language guidance in VLMbench tasks. 
(2) The Language-Only agent nearly fails at all the tasks as it is visually blind and thus unable to localize the objects in the 3D space. For ``Drawer" tasks, since language instructions can provide the level information (e.g., top, middle, and bottom) and action directions (e.g., open and close), the Language-Only agent has some chance to close the drawer by collisions.
(3) Without the language guidance, the Vision-Only agent has a significant performance degradation on all tasks. It fails completely on tasks that requires more strict pose constraints, including “Drop”, “Shape Sorter”, “Pour”, and “Door”. 
For other tasks such as Pick\&Place", "Stack," “Wipe," and “Drawer”, the Vision-Only agent can succeed on few cases (though with pretty low success rates) by randomly grasping an object in the scene for manipulation. Because those tasks have lower variance of the grasping actions and following movements.

\noindent\textbf{Ablation Study on Position and Orientation Estimation~}
We provide partial ground-truth predictions and do a unit test of the agent's position and orientation estimation abilities. The results are shown in Table~\ref{tab:results_main} and Table~\ref{tab:results_main_sub}. 
From the results, we can see the position estimation ability significantly limits the performance of 6D-CLIPort in those tasks which need correct object localization, such as ``Pick \& Place'', ``Stack'', and ``Shape Sorter'' tasks.  
One primary reason why GT position brings more improvement is that it eliminates the difficulties of localizing the target object, one of the main challenges in compositional reasoning. For example, the instruction “place the red cube into the green container” requires the model to localize the correct cube and container, and providing GT position makes the task much easier.
Besides, from the perspective of pose variances, when we divide the task into steps, the orientation of each step has fewer variances than the position in many tasks such as picking, stacking, and wiping. 
Furthermore, for the tasks requiring the cooperation of position and orientation such as ``Pour'' and ``Door'' tasks, we observe that giving partially ground truth may still not guarantee better task completion results.

More results and analysis can be found in Appendix~\ref{app:experiment}.

\begin{table}[tbp]
    \centering
    % \begin{adjustbox}{max width=\textwidth, center}
    \begin{adjustbox}{max width=\textwidth, center}{
    \begin{tabular}{lrrrrrrrr}
    \toprule
         & \multicolumn{2}{c}{\textbf{Color}} & \multicolumn{2}{c}{\textbf{Shape}}
         & \multicolumn{2}{c}{\textbf{Size}} & \multicolumn{2}{c}{\textbf{Relative Position}}
         \\
         \cmidrule(lr){2-3} \cmidrule(lr){4-5} \cmidrule(lr){6-7} \cmidrule(lr){8-9} 
         Agent & Seen & Unseen
                & Seen & Unseen
                & Seen & Unseen
                & Seen & Unseen \\
        \toprule
        Language-Only
        & 0.00 & 0.00
        & 0.00 & 0.00
        & 0.00 & 0.00
        & 0.00 & 0.00
        \\
        Vision-Only 
        & 6.00 & 5.00
        & 6.00 & 8.50
        & 6.05 & 5.10
        & 6.80 & 5.10
        \\
        6D-CLIPort
        & \textbf{15.17} & \textbf{13.00}
        & \textbf{23.00} & \textbf{19.50}
        & \textbf{18.35} & \textbf{14.92}
        & \textbf{15.31} & \textbf{15.82}
        \\
        \midrule
        6D-CLIPort (GT Ori)
        & 18.67 & 17.67
        & 28.00 & 27.25
        & 20.97 & 17.74
        & 21.43 & 18.37 \\
        6D-CLIPort (GT Pos)
        & 40.17 & 35.00
        & 56.25 & 51.25
        & 44.96 & 38.51
        & 38.61 & 34.86 \\
         \bottomrule
         & \multicolumn{2}{c}{\textbf{Direction}} & \multicolumn{2}{c}{\textbf{Level}}
        &\multicolumn{2}{c}{\textbf{Action Type}} & \multicolumn{2}{c}{\textbf{Amount}} \\
        \cmidrule(lr){2-3} \cmidrule(lr){4-5} \cmidrule(lr){6-7} \cmidrule(lr){8-9} 
             & Seen & Unseen
                & Seen & Unseen
                & Seen & Unseen
                & Seen & Unseen \\
        \toprule
        Language-Only
        & 0.00 & 0.00
        & 4.17 & 1.04
        & 2.04 & 0.51
        & 2.04 & 0.51
        \\
        Vision-Only 
        & 21.00 & 24.00
        & 14.58 & 7.29
        & 7.14 & 3.57
        & 7.14 & 3.57
        \\
        6D-CLIPort 
        & \textbf{22.00} & \textbf{26.00}
        & \textbf{22.92} & \textbf{15.63}
        & \textbf{14.29} & \textbf{10.20}
        & \textbf{14.29} & \textbf{10.20}
        \\
        \midrule
        6D-CLIPort (GT Ori)
        & 26.00 & 27.00
        & 23.96 & 17.71
        & 14.80 & 11.22
        & 14.80 & 11.22 \\
        6D-CLIPort (GT Pos)
        & 53.00 & 41.00
        & 43.75 & 52.08
        & 35.20 & 39.29
        & 35.20 & 39.29 \\
         \bottomrule
    \end{tabular}}
    \end{adjustbox}
    \caption{Success rates of all variations, including both seen and unseen settings.}
    \label{tab:results_main_sub}
    % \vspace{-0.5cm}
\end{table}

\section{Conclusion and Future Work}
\label{sec:discussion}
Vision-and-Language Manipulation (VLM) tasks are essential since they are inevitable for embodied AI. For further research in this area, we propose VLMbench, which includes various VLM tasks, and AMSolver, used for automatic VLM task generation. In addition, we test the 6D-CLIPort agent, a keypoint-based 6 DoF agent, on the benchmark. The results show that the current models can finish VLM tasks, but it is still a new area needed to be explored. We hope the VLMbench can push the research in finding general language-guided manipulation agents.

\textbf{Limitations} Our work still has limitations that can be improved by future work. First, we only consider rigid body object manipulation in the VLMbench. It is important to include the soft material objects in the future. Second, indirect manipulation tasks, like throwing the ball and playing billiards, are not included in the VLMbench. Third, since we generated data with template languages in simulator, the gap between the virtual environment and real world cannot be ignored.

\textbf{Ethical Concerns} We do not see significant risks of security threats or human rights violations in our work. Since our work contributes to the field of language-guided manipulations, we do not encourage real world robot experiments depending on our benchmark without any real world data fine-tuning. Due to the gap between the simulator and real world, the agents may execute unexpected actions.

% \textbf{Acknowledgements.} We thank Kaiwen Zhou, Yue Fan, Xuehai He, Jing Gu,and Eliana Stefani for feedback on the draft of the paper. We thank Peiqi Su for help on the object models processing.

% \clearpage

\bibliographystyle{splncs04}
\bibliography{egbib}
% \newpage
% \input{checklist}

\newpage
\appendix
% \section*{Appendices}
% \addcontentsline{toc}{section}{Appendices}
% \renewcommand{\thesubsection}{\Alph{subsection}}
Project website:\url{https://sites.google.com/ucsc.edu/vlmbench/home}
\section{Task Details}
\label{app:task}
% \begin{wraptable}{r}{0.5\textwidth}
\begin{table}[htbp]
\centering
\caption{All task variations except shape used in VLMbench. The shape variation of each task can be found in the detail descriptions of each task category.}
\label{table:variations}
% \resizebox{\textwidth}{!}{%
\begin{tabular}{ccc}
\midrule
\textbf{Variations} & \textbf{Totals} & \textbf{Values}                                                                           \\ \midrule
Color &
  25 &
  \begin{tabular}[c]{@{}c@{}}seen:red, maroon, lime, green, blue,navy, yellow, cyan, magenta, \\ silver, gray, olive, purple, teal, azure, violet, rose, black, white\\
  unseen: brown, gold, pink, chocolate, coral\end{tabular} \\ \midrule
Size                                                        & 5      & \begin{tabular}[c]{@{}c@{}}larger, smaller, large,  medium, small\end{tabular} \\ \midrule
\begin{tabular}[c]{@{}c@{}}Relative\\ Position\end{tabular} & 5      & top, front, rear, left, right                                                    \\ \midrule
% Shape &
%   13 &
%   \begin{tabular}[c]{@{}c@{}}seen: cube,  cylinder, triangular prism, rectangle,\\ circle, star, moon, door, drawer, mug \\
%   unseen: letter of t, flower, cross\end{tabular} \\ \midrule
% Direction                                                   & 2      & horizontal, vertical                                                             \\ \midrule
Level                                                       & 3      & top, middle, bottom                                                              \\ \midrule
Amount                                                      & 2      & fully, slightly                                                                  \\ \midrule
\begin{tabular}[c]{@{}c@{}}Action\\ Type\end{tabular}& 2      & open, close                                                                      \\ \midrule
\end{tabular}%
% }
% \end{wraptable}
\vspace{-0.5cm}
\end{table}

\begin{table}[htbp]
\centering
\caption{All object models used in VLMbench. The number behind the object class indicate the instance number of that class.}
\label{table:object_models}
\resizebox{\textwidth}{!}{%
\begin{tabular}{ccc}
\midrule
\textbf{Object type} & \textbf{Number of classes} & \textbf{Classes}                                                                           \\ \midrule
Basic model &
  3 &
  \begin{tabular}[c]{@{}c@{}}cube (1), triangular prism (1), cylinder (1)\end{tabular} \\ \midrule
Special model  & 9      & \begin{tabular}[c]{@{}c@{}}star (1), moon (1), cross (1), flower (1), letter of 't' (1),\\ pencil (1), basket (1), box container(1), shape sorter (1)\end{tabular} \\ \midrule
Planar model & 6      & rectangle (1), circle (1), triangle (1), star (1), cross (1), flower (1)     \\ \midrule
Functional model &
  2 &
  \begin{tabular}[c]{@{}c@{}}mug (6), sponge (1)\end{tabular} \\ \midrule
Articulated model                                                   & 2      &  \begin{tabular}[c]{@{}c@{}}door with one rotatble handle (2), \\cabinet with three vertical drawers (3)\end{tabular}                                                        \\ \midrule
\end{tabular}%
}
% \end{wraptable}
\vspace{-0.5cm}
\end{table}

In the VLMbench, we show eight task categories:``Pick \& Place objects", ``Stack objects", ``Drop pencil", ``Put into shape sorter", ``Pour water", ``Wiper table", ``Use drawer", and ``Use door". Here, we list variations used for these tasks in Table.~\ref{table:variations}. For each demonstration, all things in the scene will change the pose at the beginning. When building an instance-level task with one variation, the other variations will also randomly change. For example, in the demonstrations of ``Pick \& Place objects" with ``size" variation, all objects' color and relative positions, including targets and distractors, will randomly change. 
In the dataset, we have five types of objects, shown in Table~\ref{table:object_models}. We will explain each task in detail as follows. Visualizations can be found in the project website.

\subsection{Pick \& Place Objects}
\noindent\textbf{Task Definition:}
The agent needs to distinguish the specific object to grasp and then place it into a particular container. The object can be placed anywhere with any orientation inside the container.

% \noindent\textbf{Variations:} Color, Size, Relative Position, and Shape. These variations work on all objects and containers except the shape of containers.

\noindent\textbf{Task Templates:} Unit task sequence: \textit{(Control, target object)+(M1[a1,b1], target object)}; Goal Conditions: A 3D bounding box without orientation constraints inside the empty space of target container, for \textit{(M1[a1,b1], target object)}.

\noindent\textbf{Success Conditions:} The object detector inside the target container only can be triggered by the specific object. When the detector is triggered, the task considers a success.

\noindent\textbf{Instruction Templates:} High-level instructions: ``Pick up [target object description] and place it into [target container description]."; Low-level instructions: (``Move to the top of [target object description]; Grasp [target object description].", ``Move the object into [target container description]; Release the gripper.").

\noindent\textbf{Object models:}  One container model is used in both seen and unseen. In the seen settings, five object models: star, triangular, cylinder, cube, and moon. In the unseen settings, four object models: cube, the letter of 't,' cross, and flower.

\noindent\textbf{Variations and scene settings:} \\
All objects are randomly changing colors, size, and positions in each demonstration.\\
\textit{Color}: There are two same-shape objects and two same-shape containers in the scene initialization. All colors are randomly sampled from the color library. The object description is ``[color] object"; The container description is ``[color] container." \\
\textit{Size}: There are two same-shape objects and two same-shape containers in the scene initialization. One object and one container are randomly magnified while others are randomly shrunk. The object description is ``[larger/smaller] object"; The container description is ``[larger/smaller] container."\\
\textit{Relative Position}: There are two same-shape objects and two same-shape containers in the scene initialization. All objects are randomly sampled in the workspace until they obey the predefined relative positions. The object description is ``[front/rear/left/right] object"; The container description is ``[front/rear/left/right] container."\\
\textit{Shape}: There are two same-shape containers and more than two objects with different shapes in the scene initialization. The number of objects varies from two to the length of the object library. The object description is ``[shape]"; The container descriptions is ``[color] container."

\subsection{Stack Objects}
\noindent\textbf{Task Definition:}
The agent needs to distinguish the specific two objects and then stack them in a particular sequence. Since the objects have different shapes and some surfaces are hard to maintain for stacking, we use plane surfaces. Therefore, the goal pose of the above object should be inside the top plane of the below object and only can rotate along the axis which is perpendicular to the plane.

% \noindent\textbf{Variations:} Color, Size, Relative Position, and Shape. These variations work on all objects.

\noindent\textbf{Task Templates.} Unit task sequence: \textit{(Control, target object)+(M1[a2,b2], above object)}; Goal Conditions: A plane on the surface of the below object with orientation constraints along the perpendicular axis, for \textit{(M1[a2,b2], above object)}.

\noindent\textbf{Success Conditions.} The object detector attached to the bottom object will be triggered when the specific thing is above.

\noindent\textbf{Instruction Templates.} High-level instructions: ``Stack [below object description] and [above object description] in sequence."; Low-level instructions: (``Move to the top of [above object description]; Grasp [above object description].", ``Move the object on [below object description]; Release the gripper.").

\noindent\textbf{Object models:} In the seen settings, five object models: star, triangular, cylinder, cube, moon. In the unseen settings, four object models: cube, the letter of 't', cross, flower.

\noindent\textbf{Variations and scene settings:} \\
All objects are randomly changing colors, size, and positions in each demonstration.\\
\textit{Color}: There are four same-shape objects in the scene initialization. All colors are randomly sampled from the color library. The object descriptions are all ``[color] [shape]." \\
\textit{Size}: There are three same-shape objects in the scene initialization. One model is randomly magnified while another is randomly shrunk. The object description is ``[large/medium/small] [shape]."\\
\textit{Relative Position}: There are four objects with two different shapes in the scene initialization. All objects are randomly sampled in the workspace until the relative position of the same-shape objects obey the predefined relative positions. The below object description is ``[front/rear/left/right] [shape 1]"; The above object description is ``[front/rear/left/right] [shape 2]."\\
\textit{Shape}: There are more than two objects with different shapes in the scene initialization. The number of objects varies from two to the length of the object library. The below object description is ``[shape 1]"; The above object description is ``[shape 2]."

\subsection{Drop Pencil}
\noindent\textbf{Task Definition:}
The agent must distinguish the specific pencil and then drop it into an upright container. Only when the pencil is vertical down above the container can it fall appropriately. 

% \noindent\textbf{Variations:} Color, Size, and Relative Position. These variations work on all pencils and containers.

\noindent\textbf{Task Templates.} Unit task sequence: \textit{(Control, target pencil)+(M1[a3,b2], target pencil)}; Goal Conditions: A line, which is perpendicular to the opening of target container, with orientation constraints along the line, for \textit{(M1[a3,b2], target pencil)}.

\noindent\textbf{Success Conditions.} The object detector attached to the target container will be triggered when the specific pencil is inside.

\noindent\textbf{Instruction Templates.} High-level instructions: ``Drop [target pencil description] into [target container description]."; Low-level instructions: (``Move to the top of [target pencil description]; Grasp [target pencil description].", ``Move the object above [target container description]; Release the gripper.").

\noindent\textbf{Object models:} One pencil and one basket for both seen and unseen.

\noindent\textbf{Variations and scene settings:} \\
All objects are randomly changing colors and positions in each demonstration.\\
\textit{Color}: There are two same-shape pencils and two same-shape baskets in the scene initialization. All colors are randomly sampled from the color library. The object description is ``[color] pencil"; The container description is ``[color] container." \\
\textit{Size}: There are two same-shape pencils and two same-shape baskets in the scene initialization. One pencil and one container are randomly magnified while others are randomly shrunk. The object description is ``[larger/smaller] pencil"; The basket description is ``[larger/smaller] container."\\
\textit{Relative Position}: There are two same-shape pencils and two same-shape baskets in the scene initialization. All objects are randomly sampled in the workspace until they obey the predefined relative positions. The object description is ``[front/rear/left/right] object"; The basket description is ``[front/rear/left/right] container."

\subsection{Put Into Shape Sorter}
\noindent\textbf{Task Definition:}
There is a shape sorter in the environment, and the agent needs to distinguish the specific object and then put it through the hole of the sorter. Only when the object's shape and pose fit the hole can it fall.

\noindent\textbf{Variations:} Color, Shape, and Relative Position. These variations work on all objects except the sorter.

\noindent\textbf{Task Templates.} Unit task sequence: \textit{(Control, target pencil)+(M1[a4,b3], target object)}; Goal Conditions: A fixed pose related to the sorter for each object shape, for \textit{(M1[a4,b3], target object)}.

\noindent\textbf{Success Conditions.} The object detector attached to the sorter will be triggered when the specific object is inside.

\noindent\textbf{Instruction Templates.} High-level instructions: ``Put [target object description] through the hole of [target hole description]."; Low-level instructions: (``Move to the top of [target object description]; Grasp [target object description].", ``Move the object to the hole of [target hole description]; Release the gripper.").

\noindent\textbf{Object models:} One shape sorter container for both seen and unseen. In the seen settings, four object models: star, triangular, cylinder, and cube. In the unseen settings, five object models: star, triangular, cylinder, cube, and moon.

\noindent\textbf{Variations and scene settings:}
All objects are randomly changing colors and positions in each demonstration.\\
\textit{Color}: There are three same-shape objects and one shape sorter in the scene initialization. All colors are randomly sampled from the color library. The object description is ``[color] [shape]." \\
\textit{Relative Position}: There are two same-shape objects and one shape sorter in the scene initialization. All objects are randomly sampled in the workspace until they obey the predefined relative positions. The object description is ``[front/rear/left/right] [shape]."\\
\textit{Shape}: There are one shape sorter and more than two objects with different shapes in the scene initialization. The number of objects varies from two to the length of the object library. The object description is "[shape]."

\subsection{Pour Water}
\noindent\textbf{Task Definition:}
The agent needs to pour the water from the specific source mug into the particular container mug. Here we use 50 small particles to simulate the water.

\noindent\textbf{Variations:} Color, Size, and Relative Position. These variations work on all mugs.

\noindent\textbf{Task Templates.} Unit task sequence: \textit{(Control, source mug)+(M2[a1,b2], source mug)+(M2[a4, b4], source mug)}; Goal Conditions: For \textit{(M2[a1,b2], source mug)}, we need to keep the opening of the source mug is upward with the rotations along the axis of the opening directions. Further, the goal position of this step needs to make the horizontal distance between the geometry centers of two mugs less than the half-height of the source mug, and the vertical distance should be larger than the height of the container mug. For \textit{(M2[a4, b4], source mug)}, the source mug needs to make the opening point to the container mug. Therefore, the end-effector should follow a particular path, keeping the source mug in the same position but rotating it to the required orientation. Therefore, the goal condition is a set of poses, calculated by a function using the geometry and poses of two mugs.

\noindent\textbf{Success Conditions.} The object detector attached to the container mug will be triggered when more than half particles are inside the mug.

\noindent\textbf{Instruction Templates.} High-level instructions: ``Pour the water from [source mug description] to [container mug description]."; Low-level instructions: (``Move to the [relative position] of [source mug description]; Grasp [source mug description].", ``Move the object to the top of [container mug description] with the opening upwards.", ``Rotate [source mug description] toward [container mug description]").

\noindent\textbf{Object models:} Four different mugs in seen scenes and two different mugs in unseen scenes.

\noindent\textbf{Variations and scene settings:}
All objects are randomly changing colors and positions in each demonstration.\\
\textit{Color}: There are three same-shape mugs in the scene initialization. All colors are randomly sampled from the color library. The object description is ``[color] mug." \\
\textit{Relative Position}: There are two same-shape mugs in the scene initialization. All objects are randomly sampled in the workspace until they obey the predefined relative positions. The object description is ``[front/rear/left/right] mug."\\
\textit{Size}: There are two same-shape mugs in the scene initialization. One mug is randomly magnified while another is randomly shrunk. The object description is ``[larger/smaller] mug."\\

\subsection{Wipe Table}
\noindent\textbf{Task Definition:}
The agent needs to wipe the particular area with a sponge, which means moving the sponge from one side of the area to another with keeping it contacting with the area.

% \noindent\textbf{Variations:} Color, Size, Shape, Directions, and Relative Position. These variations are all working on all target areas.

\noindent\textbf{Task Templates.} Unit task sequence: \textit{(Control, sponge)+(M1[a4,b2], sponge)\\+(M2[a2, b2], sponge)}; Goal Conditions: For \textit{(M1[a4,b2], sponge)}, the sponge needs to move to one side of the target area with the surface face contacting the area. For \textit{(M2[a2, b2], sponge)}, the sponge needs to move from the current side to another side with keeping contact. The goal conditions are both a set of poses with the fixed position and rotations along the axis perpendicular to the table.

\noindent\textbf{Success Conditions.} We create 50 small invisible particles in the target area, and these particles will be removed if the sponge surface touches them. The successor attached to the target area will be triggered when more than half particles have been removed.

\noindent\textbf{Instruction Templates.} High-level instructions: ``Wipe [target area description] with a sponge."; Low-level instructions: (``Move to the top of the sponge; Grasp the sponge.", ``Move the object to the side of [target area description].", ``Move the object along the main direction of [target area description]").

\noindent\textbf{Object models:} One sponge for both seen and unseen. The four planes in the seen scenes: rectangle, round, star, and triangle; The two planes in the unseen scenes: cross, flower.

\noindent\textbf{Variations and scene settings:}
All objects are randomly changing colors and positions in each demonstration.\\
\textit{Color}: There are two same-shape planes and one sponge in the scene initialization. All colors are randomly sampled from the color library. The plane description is ``[color] area." \\
\textit{Size}: There are two same-shape planes and one sponge in the scene initialization. One plane is randomly magnified while another is randomly shrunk. The plane description is ``[larger/smaller] area."\\
\textit{Relative Position}: There are two same-shape planes and one sponge in the scene initialization. All planes are randomly sampled in the workspace until they obey the predefined relative positions. The object description is ``[front/rear/left/right] area."\\
\textit{Direction}: There are two same-shape directional planes and one sponge in the scene initialization. One plane is horizontal to the width of table while another is vertical. The plane description is ``[horizontal/vertical] area."\\
\textit{Shape}: There are one sponge and more than two planes with different shapes in the scene initialization. The number of planes varies from two to the length of the object library. The plane description is ``[shape] area."

\subsection{Use Drawer}
\noindent\textbf{Task Definition:}
The agent needs to figure out the exact level of the drawers to use and execute the correct action with appropriate degrees. The initial states of the drawer will change according to the action types. For example, if the task is to open the top drawer, the top drawer will be closed at first.

% \noindent\textbf{Variations:} Level, Action Type, and Amount. These variations are all working on the drawer at the same time. For example, we regard ``Fully open the top drawer." as one situation.

\noindent\textbf{Task Templates.} Unit task sequence: \textit{(Control, target drawer handle)+(M2[a3,b3], target drawer joint)}; Goal Conditions: For \textit{(M2[a3,b3], target drawer)}, the agent needs to move the target drawer to a fixed position with the linear physic constraints on the joints of the drawer. Therefore, the goal conditions are a set of the end-effector poses whose positions are along an axis, and the orientations are fixed.

\noindent\textbf{Success Conditions.} The successor will be triggered when the joint of the target drawer have moved a particular length.

\noindent\textbf{Instruction Templates.} High-level instructions: ``[Amount] [Action Type] the [Level] drawer."; Low-level instructions: (``Move to the front of the handle of [target drawer description]; Grasp the handle of [target drawer description].", ``Move along the axis of [target drawer description] for [Amount] [Action Type].").

\noindent\textbf{Object models:} Two cabinets in seen scenes and one cabinet in unseen scenes. The cabinets have different appearance but all have three vertical drawers.

\noindent\textbf{Variations and scene settings:}
The cabinet is randomly changing the pose in the workspace for each demonstration.

\textit{Level, Action Type, and Amount}: One cabinet is sampled from the object list. These variations are all working on the drawer at the same time. For example, we regard ``Fully open the top drawer." as one situation.

\subsection{Use Door}
\noindent\textbf{Task Definition:}
The agent needs to execute the correct action with appropriate degrees for the door. If the door is closed, the agent needs to rotate the door handle to unlock it for opening it. Like the use drawer tasks, the initial states will change according to the action type.

% \noindent\textbf{Variations:} Action Type and Amount. These variations are all working on the door at the same time. For example, we regard ``Fully open the door." as one situation.

\noindent\textbf{Task Templates.} Unit task sequence: For closing the door, \textit{(Control, door handle)+(M2[a4,b4],door plank joint)}. For opening the door, \textit{(Control, door handle)+(M2[a4,b4],door handle joint)+(M2[a4,b4],door plank joint)}; Goal Conditions: For \textit{(M2[a4,b4],door handle joint)}, the agent needs to rotate the joint, which connects the handle and plank, to a degree for unlock. Therefore, the goal conditions are a set of the end-effector poses whose positions are along an arc, and the orientations are depended on the positions. For \textit{((M2[a4,b4], door plank joint)}, the goal conditions are the same as the previous one, but the joint connects the plank and base.

\noindent\textbf{Success Conditions.} The successor will be triggered when the plank and base joint has reached a particular angle.

\noindent\textbf{Instruction Templates.} High-level instructions: ``[Amount] [Action Type] the door."; Low-level instructions: (``Move to the front of the handle of the door; Grasp the handle of the door.", ``Rotate around the axis of the handle joint.", ``Rotate around the axis of the door joint for [Amount] [Action Type].").

\noindent\textbf{Object models:} One door in seen scenes and one door in unseen scenes. The doors have different appearance but all have the rotatable handle.

\noindent\textbf{Variations and scene settings:}
The door is randomly changing the pose in the workspace for each demonstration.

\textit{Action Type and Amount}: One door is sampled from the object list. These variations are all working on the door at the same time. For example, we regard ``Fully open the door." as one situation.

\section{Implementation Details}
\label{app:impl}
\subsection{AMSolver}

% \subsection{Task Creation}
\noindent\textbf{Task Creation}
To generate demonstrations for semantic tasks, we need to use the predefined unit task sequence combined with object configurations and goal conditions. The goal conditions are the feasible goal pose sets of the current manipulated object for each unit task. The goal conditions can be formatted as bounding boxes, planes, lines, joint angles, or generated by functions for special pose sets. For example, a rearrangement task \textit{``Put the ball into the basket"} needs to transit a ball into the empty space of the basket without any constraints requirement, which means the goal condition of the ball is a set of poses inside the blank space of basket. Then, we can format this task as $(Control, ball) + (M1[a1, b1], ball)$. Note that \textit{ball} can be replaced with other objects, so $(Control, object)+ (M1[a1, b1], object)$ can represent any pick-and-place task without specific constraints. 
Another example is \textit{``Pour the water from one mug to another"}. This task needs to keep the mug opening upward during the move towards the top of another mug. We can format this task as $(Control, source\ mug) + (M2[a1, b2], source\ mug) + (M2[a4, b4], source\ mug)$. For the first $(M2[a1, b2], source\ mug)$, the goal condition should make the horizontal distance between the geometry centers of two mugs less than the half-height of the source mug, and the vertical distance should be larger than the height of the container mug. Therefore, the goal condition is a set of poses, calculated by a function using the geometry and poses of two mugs.

\noindent\textbf{Implementation}
The AMSolver is built inside the CoppeliaSim~\cite{6696520} environment and written in Python and Lua based on RLbench~\cite{james2020rlbench} and PyRep~\cite{james2019pyrep}. The agent is a Franka Emika Panda robot arm with 7 DoF, standing on a wooden table. To achieve the grasping ability for any object shape, we implement one version of the algorithm of ten Pas et al.~\cite{ten2017grasp}. The generator will calculate the normals and principal axis of partial point clouds. Then, the normals, principal axis, and cross-product will establish the grasp poses. Finally, the agent will calculate inverse kinematic to check the validation. More details can be found in~\cite{ten2017grasp}. Each unit task will generate one waypoint as the sub-goal of the end-effector or one fixed cartesian path as the fixed sub-trajectory for the end-effector in the environment. We use the OMPL library~\cite{sucan2012the-open-motion-planning-library} to find a valid trajectory between two waypoint configurations. In the end, we will get a valid path from the beginning to the task finish, consisting of the critical waypoints information. The object detector used in the simulator is a 3D box, where the poses (positions and orientations) and properties (length, width, height) can be modified. The object detector is placed in the target destinations of the task. When the target object mesh has overlapped with the detector, it will return the true value for checking the object.

\subsection{VLMbench}
\noindent\textbf{Dataset Generation}
We collect the VLMbench dataset in the environment of RLbench with AMSolver. There are five RGB-D cameras in the environment: front view, left view, right view, overhead view, and wrist view. We use the image resolution of $360\times360$ in this dataset. As mentioned in section 3.4, the AMSolver will output the trajectories with waypoints. Therefore, the robot arm will use motion planning to transfer from one waypoint to another, and the simulator will record the observation with the time step 50 ms. Each camera can produce an RGB-D image with point cloud and instance masks for the observation. We also record the low-level states for the robot joints, including joint velocity, joint torque, joint position, and end-effector pose. In addition, the observations record the object pose at each time step and the corresponding relationship between the frames and waypoints so that the demonstrations can be automatically divided into the sub-demonstrations with the critical frame. The objects will change the poses among different demonstrations, and the distractors will be randomly added to the environment.

For language instructions, we predefined some templates for each task category to quickly generate various language instructions by filling the object properties' descriptions, shown in Fig.~\ref{fig:tasks}. For example, the ``Pick \& Place objects" task has the template ``Pick [Object] and place it into [Container]," where [Object] denotes the target object descriptions corresponding to variations mentioned above. Meanwhile, by defining the structured language templates on the unit task, we can simultaneously generate the low-level instructions in the dataset, which structurally describe the basic actions corresponding to the frames, e.g. ``Move to [relative position] [manipulated object]; Grasp [manipulated object]" correspond to the pre-grasp action and grasp action in the Control unit task. We hope this information will be helpful for further research in this area.

% \noindent\textbf{Task Statistics}
% In total, we have 24 instance-level tasks with 231 variations, which include 120 color variations, 18 size variations, 60 relative position variations, 15 shape variations, 2 direction variations, 16 variations of the combinations of level, amount, and action type.
% The benchmark has collected 5,751 manipulation demonstrations which contain 4,601 train demonstrations and 1,150 validation demonstrations. Since the agent test in the online simulator, we have generated 1460 test settings for all instance-level tasks.

\subsection{6D-CLIPort}
In this section, we provide hyperparameter values in our 6D-CLIPort training. We have trained each agent on eight A5000 GPUs for one hour with a batch size of 16, an epoch of 15, and a learning rate of 1e-3. The pixel-wise feature maps from the attention module have one dimension. The output feature maps of the value and key module have four dimensions in each pixel to estimate 2D positions and the rotation along the perpendicular axis. Meanwhile, The output feature maps have 12 dimensions in each pixel for the other three regressors and are chunked into three parts for separate convolutions, where the convolutions of value and key feature maps are input to each regressor.

\section{Dataset Details}
\label{app:dataset}
\begin{figure}[!tbp]
\centering
\includegraphics[width=0.9\columnwidth]{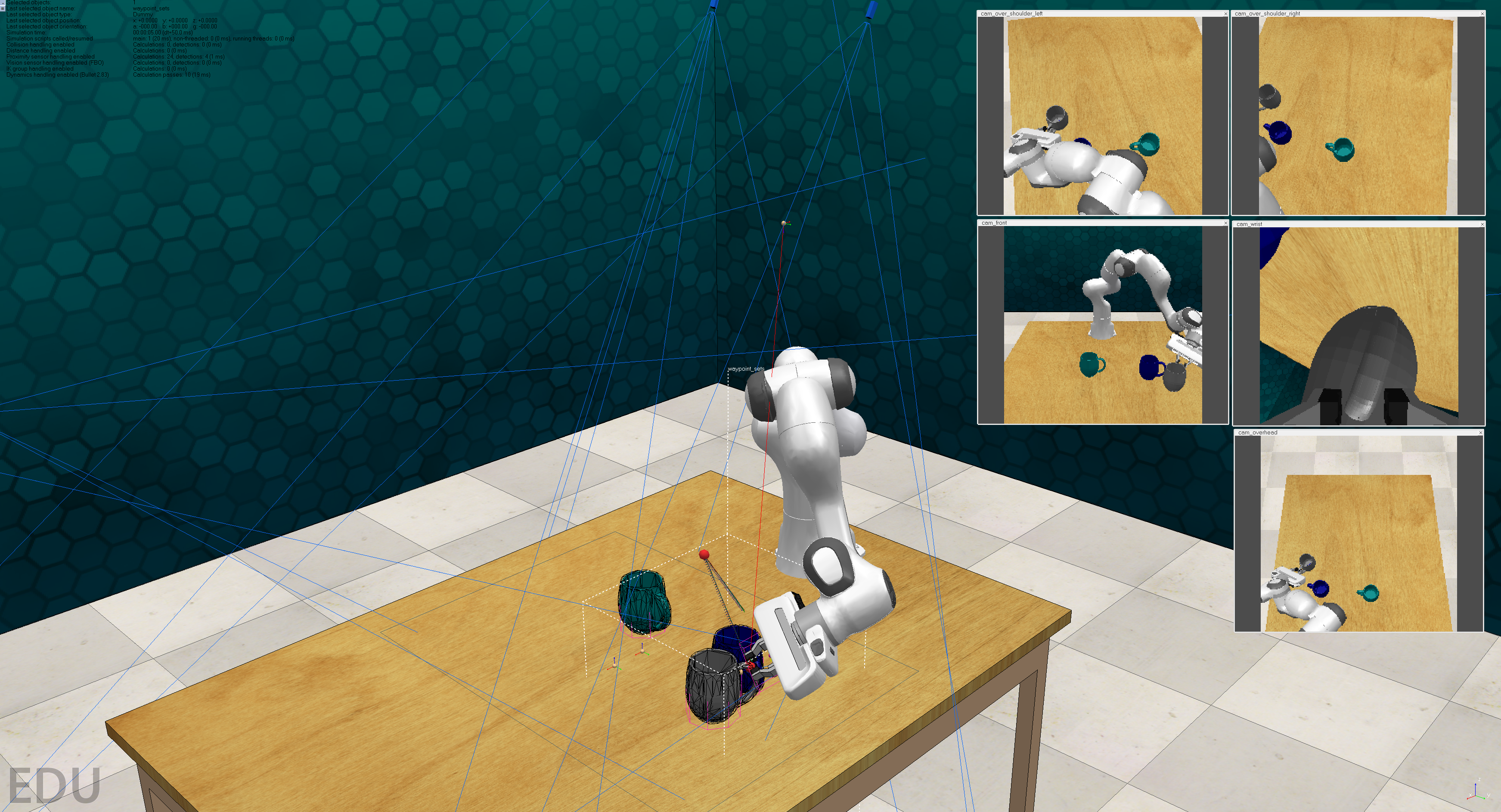}
\caption{The environment used in VLMbench. A 7 DoF robot arm stands on the table with five cameras from different views. Meanwhile, the images from the camera views are shown aside. We also can see the predefine of the paths in the environment, which are the ground truth waypoints generated by the AMSolver.}
\label{fig:env}
\end{figure}

\begin{table}[t]
    \centering
    \caption{The number of episodes for each task in the dataset.}
    \begin{adjustbox}{max width=1.2\textwidth, center}
    \begin{tabular}{clccccc}
    \toprule
         \multirow{2}{*}{Task category}&  
         \multirow{2}{*}{Variation}& 
         \multirow{2}{*}{Train} & \multicolumn{2}{c}{Valid}
         & \multicolumn{2}{c}{Test}
         \\
         \cmidrule(lr){4-5} \cmidrule(lr){6-7}
             & & & Seen & Unseen
                & Seen & Unseen \\
        \toprule
        \multirow{4}{*}{Pick} & Color & 400 & 100 & 25 & 100 & 100\\
                & Relative & 320 & 80 & 80 & 96 & 96\\
                & Shape & 100 & 25 & 20 & 100 & 100 \\
                & Size & 80 & 20 & 20 & 100 & 100 \\
        \toprule
        \multirow{4}{*}{Stack} & Color & 400 & 100 & 25 & 100 & 100\\
                & Relative & 320 & 80 & 80 & 96 & 96\\
                & Shape & 100 & 25 & 20 & 100 & 100 \\
                & Size & 120 & 30 & 30 & 96 & 96 \\
        \toprule
        \multirow{3}{*}{Drop} & Color & 400 & 100 & 25 & 100 & 100\\
                & Relative & 320 & 80 & 80 & 96 & 96\\
                & Size & 80 & 20 & 20 & 100 & 100 \\
        \toprule
        \multirow{3}{*}{Place} & Color & 400 & 100 & 25 & 100 & 100\\
                & Relative & 80 & 20 & 20 & 100 & 100\\
                & Shape & 80 & 20 & 25 & 100 & 100 \\
        \toprule
        \multirow{5}{*}{Wipe} & Color & 400 & 100 & 25 & 100 & 100\\
                & Relative & 80 & 20 & 20 & 100 & 100\\
                & Shape & 80 & 20 & 20 & 100 & 100 \\
                & Size & 40 & 10 & 10 & 100 & 100 \\
                & Direction & 40 & 10 & 10 & 100 & 100 \\
        \toprule
        \multirow{3}{*}{Pour} & Color & 400 & 100 & 25 & 100 & 100\\
                & Relative & 80 & 20 & 20 & 100 & 100\\
                & Size & 40 & 10 & 10 & 100 & 100 \\
        \toprule
        Door & action\&amount & 80 & 20 & 20 & 100 & 100\\
        \toprule
        Door & action\&amount\&level & 240 & 60 & 60 & 96 & 96\\
        \toprule
        \multicolumn{2}{c}{Total} & 4680 & 1170 & 705 & 2380 & 2380 \\
        
         \bottomrule \\
    \end{tabular}
    \end{adjustbox}
    \label{tab:all_episodes}
    \vspace{-0.5cm}
\end{table}

\begin{figure}[!tbp]
\centering
\includegraphics[width=0.9\columnwidth]{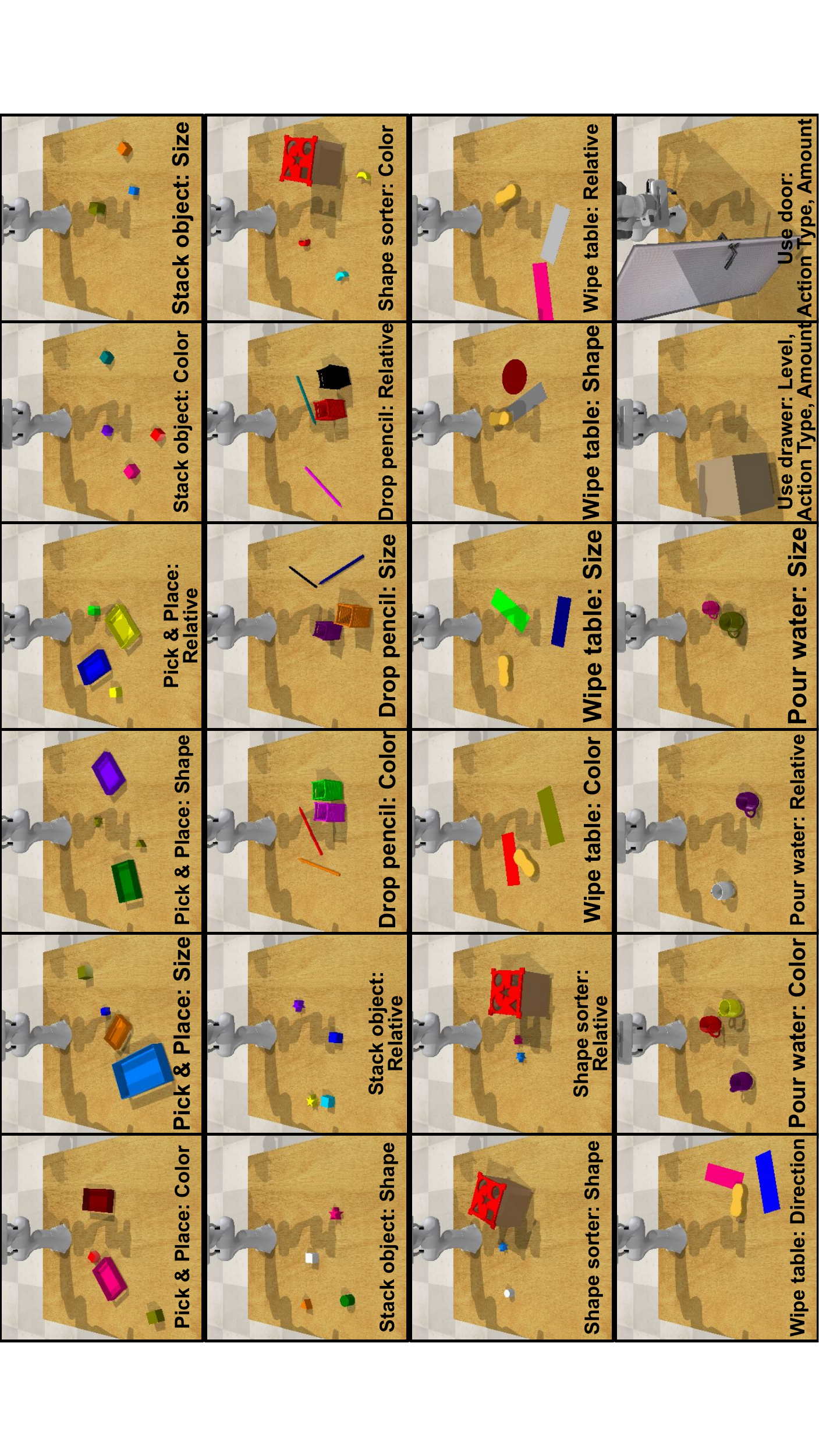}
\vspace{-1.5cm}
\caption{All instance-level tasks in VLMbench.}
\label{fig:instance}
\end{figure}

We sample VLMbench in the CoppeliaSim simulator, shown in Fig.~\ref{fig:env}. The simulator collects the data with 20 Hz frequency, and at each step, the state record all visual, robot, and objects information. To illuminate the dataset's structure, we have a folder for each instance-level task, e.g., pouring water with color variation, and each variation value creates a folder under the task folder, e.g., the target mug is red in the pouring water with color variation. Inside one variation value folder, we have demonstration folders that contain the state record for demonstrations. Then, we save the RGB-D, segmentation, and point cloud from each camera in one demonstration folder. Another file contains all low-level robot states, object states, and waypoint information for every step. All instance-level tasks are shown in Fig.~\ref{fig:instance}. Demonstrations demos can be found in the video on the website.

\noindent\textbf{Task Statistics}
In total, we have 24 instance-level tasks with 234 variations, which include 120 color variations, 18 size variations, 60 relative position variations, 18 shape variations, 2 direction variations, 16 variations of the combinations of level, amount, and action type.
The benchmark have 6,555 manipulation demonstrations which contain 4,680 train demonstrations, 1,170 seen validation demonstrations, and 705 unseen validation demonstrations. Since the agent test in the online simulator, we have generated 2,380 seen test settings and 2,380 unseen test settings for all instance-level tasks. The detail numbers can be found in Table~\ref{tab:all_episodes}. All demonstrations are generated by two servers with 64-Core CPUs and 8 GPUs within 24 hours.

\section{Additional Experiments}
\label{app:experiment}

\begin{table}[tbp]
    \centering
    \begin{adjustbox}{max width=\textwidth, center}
    \begin{tabular}{lrrrrrrrrrrrrrrrrrrrrrrrr}
    \toprule
         & \multicolumn{6}{c}{\textbf{Pick\&Place}} & \multicolumn{6}{c}{\textbf{Stack}}
         & \multicolumn{6}{c}{\textbf{Drop}} & \multicolumn{6}{c}{\textbf{Shape Sorter}}
         \\
         \cmidrule(lr){2-7} \cmidrule(lr){8-13} \cmidrule(lr){14-19} \cmidrule(lr){20-25} 
         Agent & \multicolumn{3}{c}{Seen} & \multicolumn{3}{c}{Unseen}
                & \multicolumn{3}{c}{Seen} & \multicolumn{3}{c}{Unseen}
                & \multicolumn{3}{c}{Seen} & \multicolumn{3}{c}{Unseen}
                & \multicolumn{3}{c}{Seen} & \multicolumn{3}{c}{Unseen} \\
                \cmidrule(lr){2-4} \cmidrule(lr){5-7} \cmidrule(lr){8-10}
                \cmidrule(lr){11-13} \cmidrule(lr){14-16} \cmidrule(lr){17-19}
                \cmidrule(lr){20-22} \cmidrule(lr){23-25}
                & \multicolumn{1}{c}{G-G} & \multicolumn{1}{c}{G-M} & \multicolumn{1}{c}{SC}
                & \multicolumn{1}{c}{G-G} & \multicolumn{1}{c}{G-M} & \multicolumn{1}{c}{SC}
                & \multicolumn{1}{c}{G-G} & \multicolumn{1}{c}{G-M} & \multicolumn{1}{c}{SC}
                & \multicolumn{1}{c}{G-G} & \multicolumn{1}{c}{G-M} & \multicolumn{1}{c}{SC}
                & \multicolumn{1}{c}{G-G} & \multicolumn{1}{c}{G-M} & \multicolumn{1}{c}{SC}
                & \multicolumn{1}{c}{G-G} & \multicolumn{1}{c}{G-M} & \multicolumn{1}{c}{SC}
                & \multicolumn{1}{c}{G-G} & \multicolumn{1}{c}{G-M} & \multicolumn{1}{c}{SC}
                & \multicolumn{1}{c}{G-G} & \multicolumn{1}{c}{G-M} & \multicolumn{1}{c}{SC}\\
        \toprule
        Language-Only
        & 0.00 & 0.00 & 0.00 & 0.00 & 0.00 & 0.00
        & 0.00 & 0.00 & 0.00 & 0.00 & 0.00 & 0.00
        & 0.00 & 0.00 & 0.00 & 0.00 & 0.00 & 0.00
        & 0.00 & 0.00 & 0.00 & 0.00 & 0.00 & 0.00
        \\
        Vision-Only 
        & 29.55 & 29.80 & 6.31 & 36.62 & 26.52 & 9.85
        & 34.95 & 32.65 & 6.89 & 32.65 & 22.45 & 1.79
        & 10.47 & 0.00 & 0.00 & 16.89 & 0.00 & 0.00
        & 8.33 & 8.67 & 0.00 & 4.00 & 6.33 & 0.33
        % \\
        % Image-Language
        % & \textbf{58.59} & \textbf{45.45} & \textbf{29.80} & \textbf{68.18} & 36.62 & 27.02
        % & 50.26 & 34.69 & 16.07 & 27.55 & 23.47 & 8.16
        % & 53.04 & 0.34 & 0.00 & 54.73 & 0.00 & 0.00
        % & 69.33 & \textbf{24.33} & 15.33 & \textbf{69.00} & \textbf{21.67} & \textbf{13.33}
        \\
        6D-CLIPort 
        & \textbf{58.33} & \textbf{43.33} & \textbf{28.28} & \textbf{63.14} & \textbf{41.16} & \textbf{27.53} 
        & \textbf{53.06} & \textbf{44.64} & \textbf{22.19} & \textbf{48.21} & \textbf{28.06} & \textbf{18.37} 
        & \textbf{70.61} & \textbf{9.12} & \textbf{6.42} & \textbf{65.88} & \textbf{8.78} & \textbf{6.42}
        & \textbf{73.00} & \textbf{23.67} & \textbf{17.33} & \textbf{66.67} & \textbf{18.67} & \textbf{12.33}
        \\
        \midrule
        6D-CLIPort (GT Ori)
        & 60.61 & 42.93 & 28.03 & 65.40 & 36.87 & 26.26
        & 56.38 & 44.64 & 26.53 & 51.02 & 48.98 & 26.02
        & 65.88 & 31.08 & 17.91 & 62.16 & 28.72 & 16.22
        & 72.67 & 32.67 & 24.00 & 67.33 & 24.00 & 15.67 \\
        6D-CLIPort (GT Pos)
        & 94.95 & 88.64 & 83.84 & 95.96 & 83.33 & 75.25
        & 96.94 & 69.39 & 58.93 & 96.43 & 47.70 & 50.51
        & 95.61 & 13.85 & 16.89 & 94.26 & 13.18 & 11.82
        & 92.00 & 19.33 & 18.00 & 94.00 & 22.67 & 17.33 \\
         \bottomrule \\
         & \multicolumn{6}{c}{\textbf{Pour}} & \multicolumn{6}{c}{\textbf{Wipe}}
        &\multicolumn{6}{c}{\textbf{Door}} & \multicolumn{6}{c}{\textbf{Drawer}} \\
        \cmidrule(lr){2-7} \cmidrule(lr){8-13} \cmidrule(lr){14-19} \cmidrule(lr){20-25}
          & \multicolumn{3}{c}{Seen} & \multicolumn{3}{c}{Unseen}
                & \multicolumn{3}{c}{Seen} & \multicolumn{3}{c}{Unseen}
                & \multicolumn{3}{c}{Seen} & \multicolumn{3}{c}{Unseen}
                & \multicolumn{3}{c}{Seen} & \multicolumn{3}{c}{Unseen} \\
                \cmidrule(lr){2-4} \cmidrule(lr){5-7} \cmidrule(lr){8-10}
                \cmidrule(lr){11-13} \cmidrule(lr){14-16} \cmidrule(lr){17-19}
                \cmidrule(lr){20-22} \cmidrule(lr){23-25}
                & \multicolumn{1}{c}{G-G} & \multicolumn{1}{c}{G-M} & \multicolumn{1}{c}{SC}
                & \multicolumn{1}{c}{G-G} & \multicolumn{1}{c}{G-M} & \multicolumn{1}{c}{SC}
                & \multicolumn{1}{c}{G-G} & \multicolumn{1}{c}{G-M} & \multicolumn{1}{c}{SC}
                & \multicolumn{1}{c}{G-G} & \multicolumn{1}{c}{G-M} & \multicolumn{1}{c}{SC}
                & \multicolumn{1}{c}{G-G} & \multicolumn{1}{c}{G-M} & \multicolumn{1}{c}{SC}
                & \multicolumn{1}{c}{G-G} & \multicolumn{1}{c}{G-M} & \multicolumn{1}{c}{SC}
                & \multicolumn{1}{c}{G-G} & \multicolumn{1}{c}{G-M} & \multicolumn{1}{c}{SC}
                & \multicolumn{1}{c}{G-G} & \multicolumn{1}{c}{G-M} & \multicolumn{1}{c}{SC}\\
        \toprule
        Language-Only
        & 0.00 & 0.76 & 0.00 & 0.00 & 0.00 & 0.00
        & 0.00 & 0.00 & 0.00 & 0.00 & 0.00 & 0.00
        & 0.00 & 0.00 & 0.00 & 0.00 & 0.00 & 0.00
        & 5.21 & 15.63 & 4.17 & 3.13 & 13.54 & 1.04
        \\
        Vision-Only 
        & 2.67 & 2.00 & 0.00 & 4.00 & \textbf{1.33} & 0.00
        & 99.20 & 19.40 & 19.80 & \textbf{96.00} & 21.20 & 20.80
        & 0.00 & 0.00 & 0.00 & 0.00 & 0.00 & 0.00
        & \textbf{26.04} & 21.88 & 14.58 & 1.04 & 18.75 & 7.29
        % \\
        % Image-Language
        % & \textbf{72.33} & 1.67 & 0.67 & \textbf{76.33} & \textbf{1.67} & \textbf{1.67}
        % & 93.00 & 10.00 & 8.80 & 93.80 & 8.20 & 6.60
        % & 13.00 & \textbf{21.00} & 1.00 & 0.00 & \textbf{21.00} & 0.00
        % & 2.08 & 9.38 & 1.04 & 5.21 & 14.58 & 5.21
        \\
        6D-CLIPort 
        & \textbf{61.00} & \textbf{2.00} & \textbf{1.00} & \textbf{69.67} & 1.00 & \textbf{1.00}
        & \textbf{99.20} & \textbf{21.60} & \textbf{22.40} & 95.20 & \textbf{21.60} & \textbf{21.80}
        & \textbf{29.00} & \textbf{14.00} & \textbf{6.00} & \textbf{5.00} & \textbf{15.00} & \textbf{5.00}
        & 22.92 & \textbf{25.00} & \textbf{22.92} & \textbf{8.33} & \textbf{21.88} & \textbf{15.63}
        \\
        \midrule
        6D-CLIPort (GT Ori)
        & 70.00 & 4.00 & 3.67 & 76.67 & 3.67 & 3.67
        & 99.60 & 25.40 & 25.80 & 97.20 & 25.20 & 25.20
        & 49.00 & 14.00 & 6.00 & 5.00 & 15.00 & 5.00
        & 31.25 & 30.00 & 23.96 & 8.33 & 23.96 & 17.71 \\
        6D-CLIPort (GT Pos)
        & 61.67 & 3.00 & 0.33 & 67.33 & 0.67 & 0.67
        & 100.00 & 61.00 & 60.20 & 99.40 & 57.40 & 53.40
        & 80.00 & 46.00 & 27.00 & 83.00 & 49.00 & 27.00
        & 79.17 & 43.75 & 43.75 & 87.50 & 50.00 & 52.08 \\
         \bottomrule
    \end{tabular}
    \end{adjustbox}
    \caption{Results of all tasks, including both seen and unseen settings. In the table, G-G denotes Goal-conditioned Grasp success rate, which reflect whether the agent can correctly grasp the target. G-M denotes Goal-conditioned Movement success rate, which reflect whether the agent can finish the task by using the pre-generated grasping poses. SC denotes success rates for the whole estimation trajectory. More explanations can be found in section~\ref{sec:goal_conditioned}.}
    \label{tab:results_full}
    \bigskip
    
    \begin{adjustbox}{max width=\textwidth, center}
    \begin{tabular}{lrrrrrrrrrrrrrrrrrrrrrrrr}
    \toprule
         & \multicolumn{6}{c}{\textbf{Color}} & \multicolumn{6}{c}{\textbf{Shape}}
         & \multicolumn{6}{c}{\textbf{Size}} & \multicolumn{6}{c}{\textbf{Relative Position}}
         \\
         \cmidrule(lr){2-7} \cmidrule(lr){8-13} \cmidrule(lr){14-19} \cmidrule(lr){20-25} 
         Agent & \multicolumn{3}{c}{Seen} & \multicolumn{3}{c}{Unseen}
                & \multicolumn{3}{c}{Seen} & \multicolumn{3}{c}{Unseen}
                & \multicolumn{3}{c}{Seen} & \multicolumn{3}{c}{Unseen}
                & \multicolumn{3}{c}{Seen} & \multicolumn{3}{c}{Unseen} \\
                \cmidrule(lr){2-4} \cmidrule(lr){5-7} \cmidrule(lr){8-10}
                \cmidrule(lr){11-13} \cmidrule(lr){14-16} \cmidrule(lr){17-19}
                \cmidrule(lr){20-22} \cmidrule(lr){23-25}
                & \multicolumn{1}{c}{G-G} & \multicolumn{1}{c}{G-M} & \multicolumn{1}{c}{SC}
                & \multicolumn{1}{c}{G-G} & \multicolumn{1}{c}{G-M} & \multicolumn{1}{c}{SC}
                & \multicolumn{1}{c}{G-G} & \multicolumn{1}{c}{G-M} & \multicolumn{1}{c}{SC}
                & \multicolumn{1}{c}{G-G} & \multicolumn{1}{c}{G-M} & \multicolumn{1}{c}{SC}
                & \multicolumn{1}{c}{G-G} & \multicolumn{1}{c}{G-M} & \multicolumn{1}{c}{SC}
                & \multicolumn{1}{c}{G-G} & \multicolumn{1}{c}{G-M} & \multicolumn{1}{c}{SC}
                & \multicolumn{1}{c}{G-G} & \multicolumn{1}{c}{G-M} & \multicolumn{1}{c}{SC}
                & \multicolumn{1}{c}{G-G} & \multicolumn{1}{c}{G-M} & \multicolumn{1}{c}{SC}\\
        \toprule
        Language-Only
        & 0.00 & 0.33 & 0.00 & 0.00 & 0.00 & 0.00
        & 0.00 & 0.25 & 0.00 & 0.00 & 0.00 & 0.00
        & 0.00 & 0.00 & 0.00 & 0.00 & 0.00 & 0.00
        & 0.00 & 0.00 & 0.00 & 0.00 & 0.00 & 0.00
        \\
        Vision-Only 
        & 29.67 & 14.17 & 6.00 & 30.00 & 13.00 & 5.00
        & 42.75 & 21.00 & 6.00 & 46.25 & 20.00 & 8.50
        & 36.49 & 15.12 & 6.05 & 36.09 & 12.76 & 5.10
        & 31.46 & 19.73 & 6.80 & 31.97 & 12.76 & 5.10
        % \\
        % Image-Language
        % & 65.00 & 17.00 & 9.33 & 62.33 & 14.50 & 8.00
        % & \textbf{71.50} & 29.25 & 20.75 & 68.00 & 20.50 & 14.00
        % & 65.52 & 17.54 & 9.88 & \textbf{65.52} & 14.52 & 8.67
        % & 64.80 & 21.60 & 12.24 & 64.63 & 16.16 & 10.03
        \\
        6D-CLIPort
        & \textbf{69.83} & \textbf{19.67} & \textbf{15.17} & \textbf{69.00} & \textbf{17.50} & \textbf{13.00}
        & \textbf{69.75} & \textbf{33.25} & \textbf{23.00} & \textbf{71.25} & \textbf{27.25} & \textbf{19.50}
        & \textbf{70.56} & \textbf{25.40} & \textbf{18.35} & \textbf{64.31} & \textbf{21.77} & \textbf{14.92}
        & \textbf{68.03} & \textbf{27.72} & \textbf{15.31} & \textbf{69.56} & \textbf{20.75} & \textbf{15.82}
        \\
        \midrule
        6D-CLIPort (GT Ori)
        & 72.00 & 24.67 & 18.67 & 70.83 & 26.00 & 17.67
        & 70.25 & 39.75 & 28.00 & 72.50 & 33.75 & 27.25
        & 71.57 & 29.23 & 20.97 & 66.53 & 28.02 & 17.74
        & 70.58 & 33.33 & 21.43 & 71.09 & 29.42 & 18.37 \\
        6D-CLIPort (GT Pos)
        & 90.67 & 42.33 & 40.17 & 90.33 & 37.50 & 35.00
        & 96.00 & 59.75 & 56.25 & 96.75 & 53.25 & 51.25
        & 90.32 & 47.58 & 44.96 & 91.53 & 41.33 & 38.51
        & 89.12 & 41.50 & 38.61 & 90.99 & 36.39 & 34.86 \\
         \bottomrule \\
         & \multicolumn{6}{c}{\textbf{Direction}} & \multicolumn{6}{c}{\textbf{Level}}
        &\multicolumn{6}{c}{\textbf{Action Type}} & \multicolumn{6}{c}{\textbf{Amount}} \\
        \cmidrule(lr){2-7} \cmidrule(lr){8-13} \cmidrule(lr){14-19} \cmidrule(lr){20-25}
          & \multicolumn{3}{c}{Seen} & \multicolumn{3}{c}{Unseen}
                & \multicolumn{3}{c}{Seen} & \multicolumn{3}{c}{Unseen}
                & \multicolumn{3}{c}{Seen} & \multicolumn{3}{c}{Unseen}
                & \multicolumn{3}{c}{Seen} & \multicolumn{3}{c}{Unseen} \\
                \cmidrule(lr){2-4} \cmidrule(lr){5-7} \cmidrule(lr){8-10}
                \cmidrule(lr){11-13} \cmidrule(lr){14-16} \cmidrule(lr){17-19}
                \cmidrule(lr){20-22} \cmidrule(lr){23-25}
                & \multicolumn{1}{c}{G-G} & \multicolumn{1}{c}{G-M} & \multicolumn{1}{c}{SC}
                & \multicolumn{1}{c}{G-G} & \multicolumn{1}{c}{G-M} & \multicolumn{1}{c}{SC}
                & \multicolumn{1}{c}{G-G} & \multicolumn{1}{c}{G-M} & \multicolumn{1}{c}{SC}
                & \multicolumn{1}{c}{G-G} & \multicolumn{1}{c}{G-M} & \multicolumn{1}{c}{SC}
                & \multicolumn{1}{c}{G-G} & \multicolumn{1}{c}{G-M} & \multicolumn{1}{c}{SC}
                & \multicolumn{1}{c}{G-G} & \multicolumn{1}{c}{G-M} & \multicolumn{1}{c}{SC}
                & \multicolumn{1}{c}{G-G} & \multicolumn{1}{c}{G-M} & \multicolumn{1}{c}{SC}
                & \multicolumn{1}{c}{G-G} & \multicolumn{1}{c}{G-M} & \multicolumn{1}{c}{SC}\\
        \toprule
        Language-Only
        & 0.00 & 0.00 & 0.00 & 0.00 & 0.00 & 0.00
        & 5.21 & 15.63 & 4.17 & 3.13 & 13.54 & 1.04
        & 2.55 & 7.65 & 2.04 & 1.53 & 6.63 & 0.51
        & 2.55 & 7.65 & 2.04 & 1.53 & 6.63 & 0.51
        \\
        Vision-Only 
        & \textbf{99.00} & 15.00 & 21.00 & 95.00 & 22.00 & 24.00
        & 26.04 & 21.88 & 14.58 & 1.04 & 18.75 & 7.29
        & 12.76 & 10.71 & 7.14 & 0.51 & 9.18 & 3.57
        & 12.76 & 10.71 & 7.14 & 0.51 & 9.18 & 3.57
        % \\
        % Image-Language
        % & 94.00 & 12.00 & 11.00 & 94.00 & 10.00 & 9.00
        % & 2.08 & 9.38 & 1.04 & 5.21 & 14.58 & 5.21
        % & 7.65 & 15.31 & 1.02 & 2.55 & 17.86 & 2.55
        % & 7.65 & 15.31 & 1.02 & 2.55 & 17.86 & 2.55
        \\
        6D-CLIPort 
        & 98.00 & \textbf{19.00} & \textbf{21.00} & \textbf{100.00} & \textbf{22.00} & \textbf{26.00}
        & \textbf{22.92} & \textbf{25.00} & \textbf{22.92} & \textbf{8.33} & \textbf{21.88} & \textbf{15.63}
        & \textbf{26.02} & \textbf{19.39} & \textbf{14.29} & \textbf{6.63} & \textbf{18.37} & \textbf{10.20}
        & \textbf{26.02} & \textbf{19.39} & \textbf{14.29} & \textbf{6.63} & \textbf{18.37} & \textbf{10.20}
        \\
        \midrule
        6D-CLIPort (GT Ori)
        & 99.00 & 26.00 & 26.00 & 98.00 & 29.00 & 27.00
        & 31.25 & 30.00 & 23.96 & 8.33 & 23.96 & 17.71
        & 40.31 & 22.45 & 14.80 & 6.63 & 19.39 & 11.22
        & 40.31 & 22.45 & 14.80 & 6.63 & 19.39 & 11.22 \\
        6D-CLIPort (GT Pos)
        & 100.00 & 63.00 & 53.00 & 100.00 & 56.00 & 41.00
        & 79.17 & 43.75 & 43.75 & 87.50 & 50.00 & 52.08
        & 79.59 & 44.90 & 35.20 & 85.20 & 49.49 & 39.29
        & 79.59 & 44.90 & 35.20 & 85.20 & 49.49 & 39.29 \\
         \bottomrule
    \end{tabular}
    \end{adjustbox}
    \caption{Results of all variations. The notations are used the same as in Table~\ref{tab:results_main}.}
    \label{tab:results_full_sub}
    
\end{table}

We have shown task success rates of each baseline agent in both seen and unseen settings. Here, we provide the results of step success for both seen and unseen settings. Then, we want to provide more explanation and analysis of failure cases and ablations study.

\subsection{Goal-Conditioned Success Rates}
\label{sec:goal_conditioned}

In Table~\ref{tab:results_full} and~\ref{tab:results_full_sub}, we show the success rates of each step for every agent. For tasks in the VLMbench, each task can be considered as a multi-step task, since each task at least consists of one grasping step and one following movement step. Therefore, we used other two metrics for the step’s success: Goal-condition Grasp (G-G) success rate and Goal-conditioned Movement (G-M) success rate. The Goal-conditioned Grasp (G-G) success rates indicate whether the agent has grasped the correct object for the first step. The Goal-conditioned Movement (G-M) success rates indicate whether the agent can satisfy the success conditions in the following movement by using the pre-generated grasp poses for the grasping step. We can find that the agent has high grasping successes in both seen and unseen settings for most tasks, including pick, stack, drop, shape sorter, pour, and wiper, which indicates that 6D-CLIPort can successfully reason the correct grasping objects with both seen and unseen settings. For these tasks, the prominent failure cases come from the following movement, where the conclusion also can be gotten from the minor difference between the goal-conditioned movement and total success rate. In the wiping tasks, since we only have one sponge shape model in both training and testing, we can see a high grasping success rates on all settings. We also can find that the Vision-Only agent has a higher grasping success rate than 6D-CLIPort in the ``Drawer" tasks with seen settings, but it has a much lower grasping success rate in the unseen settings, which indicates that the Vision-Only agent has overfitted on the grasping step in the ``Drawer" tasks. For opening/closing drawer and door tasks, we find that goal-conditioned movement success rates are similar for seen and unseen settings, meaning the agent performs similarly after the agent has correct grasping. In opening/closing drawer tasks, we also find that the goal-condition grasp is smaller than total success, which means some tasks reach the success conditions without grasping the handle of doors. It also makes sense that closing the drawer can be finished by pushing instead of grasping the handle first.

\subsection{Failure Cases}
\label{failure_ref}

\begin{figure}[!tbp]
\centering
\includegraphics[width=\columnwidth]{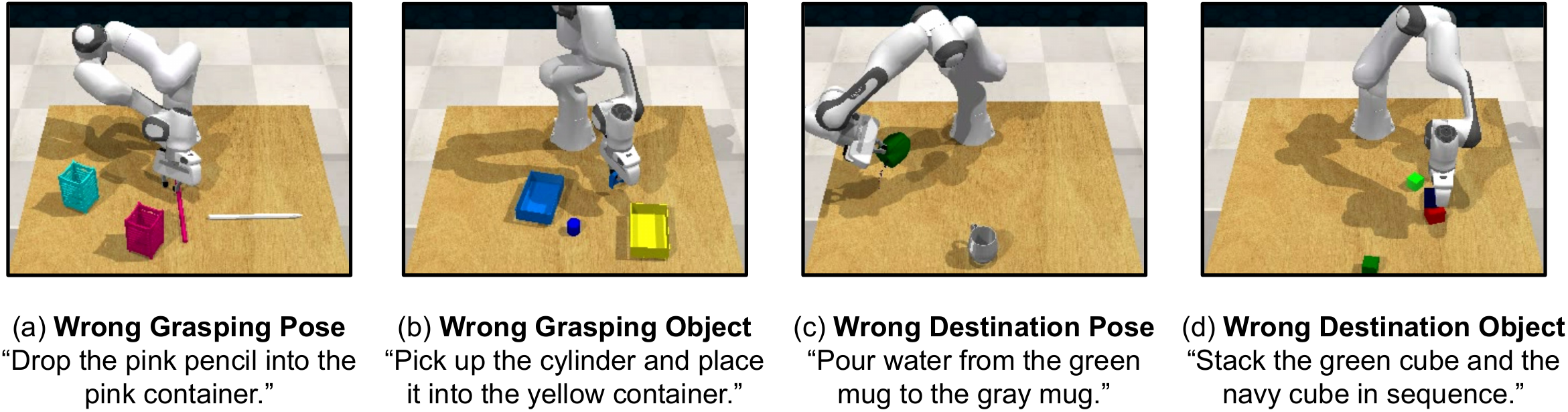}
% \vspace{-1.5cm}
\caption{Four failure cases in the results of the 6D-CLIPort. Details can be found in~\ref{failure_ref}.}
\label{fig:failure}
\end{figure}

The results show that the 6D-CLIPort agent has failed in many tasks. Except the unreachable poses due to the errors in the position and orientation estimations, we summarize the four kinds of failure cases, shown in Fig.~\ref{fig:failure}: (a) Wrong grasping pose, which means the agent cannot grasp any object. (b) Wrong grasping object, which means the agent grasps the incorrect object. (c) Wrong destination pose, which means the destination pose of the agent cannot finish any semantic task. (d) Wrong destination object, which means the estimation destination is inconsistent with the instructions. More visualization can be found in the attached video.

\begin{figure}[!tbp]
\centering
% \vspace{-1.5cm}
\includegraphics[width=0.9\columnwidth]{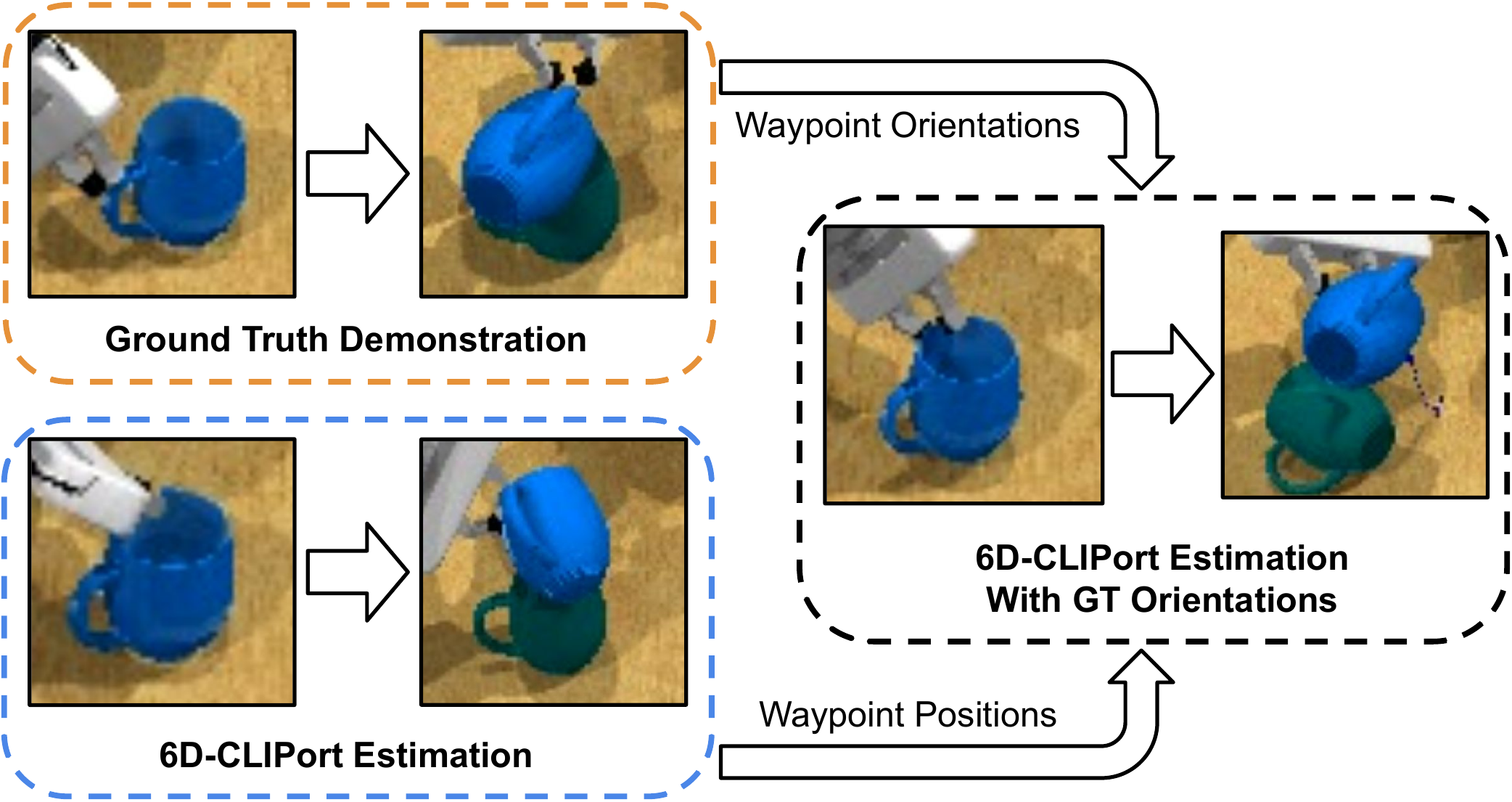}
\caption{6D-CLIPort with ground truth waypoint information. The instruction here is ``Pour water from the blue mug to the green mug." Since the orientations and positions are unmatched, we can see 6D-CLIPort with ground truth orientation can pour the water but have collisions with the container mug. In this situation, 6D-CLIPort with ground truth positions cannot find a valid path by motion planning.}
\label{fig:gt_demo}
\end{figure}

\subsection{Partially Modal Agents}
In the experiment section, we have tested the 6D-CLIPort with the ground truth waypoints' positions or orientations. 
% The results show that the agent with ground truth information performs worse than the original agent in some cases, e.g., 6D-CLIPort with ground truth positions in the pouring tasks. 
We can see the agent still fail the tasks even with this privileged information. The reason is the unmatched positions and orientations. The ground truth positions and orientations are obtained from the pre-generated waypoints. Therefore, this ground truth information is not the optimal parameter associated with the estimation. The unmatched positions and orientations can lead to a failure or even an unreachable pose where the motion planner cannot find a feasible path. For the tasks that need the cooperation of position and orientation such as dropping, pouring, and opening the door, we can figure out that giving partially ground truth still cannot finish the task correctly. We shown a example in Fig~\ref{fig:gt_demo}.Since the estimation trajectory and ground truth trajectory can be valid but different solutions for the same task. The combination of the positions and orientations will fail the task. Furthermore, the wrong estimation of positions can lead to task failure even given the ground truth orientation since the tasks in the VLMbench require correct compositional reasoning. More visualization can be found in the video on the project web page.

\end{document}